\documentclass[sigconf]{acmart}
\AtBeginDocument{%
  }

\usepackage{hyperref}
\usepackage{array}
\usepackage{booktabs}
\usepackage{multirow}
\usepackage{makecell}
\usepackage{tabularx}
\usepackage{enumitem} 
\usepackage{svg}
\usepackage{verbatim}
\usepackage{subcaption}
\usepackage{pifont}
\usepackage{adjustbox}
\usepackage{geometry}
\usepackage{ragged2e}   
\usepackage{xcolor} 
\usepackage{wrapfig}

\usepackage{wrapfig}
\usepackage{booktabs}
\usepackage{multirow}
\usepackage{xcolor}

\newcolumntype{Y}{>{\Centering\arraybackslash}X}  

\setcopyright{none}             
\renewcommand\footnotetextcopyrightpermission[1]{} 
\acmConference[KDD '26]{32nd SIGKDD Conference on Knowledge Discovery and Data Mining}{August 09--13, 2026}{Jeju, Korea}

\usepackage{booktabs}  
\usepackage{tabularx}  
\usepackage{caption}   
\usepackage{multirow} 
\usepackage{amsmath}  
\usepackage{xcolor}   

\usepackage[table]{xcolor}

\usepackage{amssymb} 
\usepackage{graphicx}   

\definecolor{mygreen}{RGB}{0, 150, 0}
\definecolor{myred}{RGB}{200, 0, 0}
\newcommand{\up}[1]{\textcolor{mygreen}{\textbf{#1} $\uparrow$}}
\newcommand{\down}[1]{\textcolor{myred}{\textbf{#1} $\downarrow$}}
\newcommand{\neu}[1]{#1} 

\begin{document}

\title{Towards Event-Aware Forecasting in DeFi: Insights from On-chain Automated Market Maker Protocols}

\author{Huaiyu Jia}
\authornote{These authors contributed equally to this work.}
\email{hjia351@connect.hkust-gz.edu.cn}
\affiliation{%
  \institution{HKUST(GZ)}
  \city{GUANGZHOU}
  \country{China}
}

\author{Jiehshun You}
\authornotemark[1]
\email{24041917g@connect.polyu.hk}
\affiliation{%
  \institution{Hong Kong Polytechnic University}
  \city{Hong Kong}
  \country{Hong Kong SAR China}
}

\author{Yizhi Luo}
\email{yluo198@connect.hkust-gz.edu.cn}
\affiliation{%
  \institution{HKUST(GZ)}
  \city{GUANGZHOU}
  \country{China}
}

\author{Jingyu Liu}
\email{jliu514@connect.hkust-gz.edu.cn}
\affiliation{%
  \institution{HKUST(GZ)}
  \city{GUANGZHOU}
  \country{China}
}

\author{Shuo Sun}
\email{shuosun@hkust-gz.edu.cn}
\affiliation{%
  \institution{HKUST(GZ)}
  \city{GUANGZHOU}
  \country{China}
}

\renewcommand{\shortauthors}{Trovato et al.}

\begin{abstract}
Automated Market Makers (AMMs), as a core infrastructure of decentralized finance (DeFi), uniquely drive on-chain asset pricing through a deterministic reserve ratio mechanism. Unlike traditional markets, AMM price dynamics is triggered largely by on-chain events (e.g., swap) that change the reserve ratio, rather than by continuous responses to off-chain information. This makes event-level analysis crucial for understanding price formation mechanisms in AMMs. However, existing research generally neglects the micro-structural dynamics at the AMMs level, lacking both a comprehensive dataset covering multiple protocols with fine-grained event classification and an effective framework for event-aware modeling. To fill this gap, we construct a dataset containing 8.9 million on-chain event records from four representative AMMs protocols: Pendle, Uniswap v3, Aave and Morpho, with precise annotations of transaction type and block height timestamps. Furthermore, we propose an Uncertainty Weighted Mean Squared Error (UWM) loss function, which incorporates the block interval regression term into the traditional Time-Point Process (TPP) objective function by weighting the uncertainty with homoscedasticity. Extensive experiments on eight advanced TPP architectures demonstrate that this loss function reduces the time prediction error by an average of 56.41\% while maintaining the accuracy of event type prediction, establishing a robust benchmark for event-aware prediction in the AMMs ecosystem. This work provides the necessary data foundation and methodological framework for modeling the discreteness and event-driven characteristics of on-chain price discovery. All datasets and source code are publicly available. \footnote{
\url{https://github.com/yosen-king/Deep-AMM-Events}}
\end{abstract}




\maketitle

\section{INTRODUCTION}

\begin{table}
    \centering
    \caption{Evolution of AMM State and Price Impact}
    \vspace{-0.1cm}
    \label{tab:amm_evolution}
    
    \resizebox{\columnwidth}{!}{%
        \renewcommand{\arraystretch}{1.3} 
        \begin{tabular}{l c c c c c c}
            \toprule
            \multirow{2}{*}{\textbf{Event Phase}} & \textbf{Block Height} & \multirow{2}{*}{\textbf{Action}} & \multicolumn{2}{c}{\textbf{Pool Reserves}} & \multicolumn{2}{c}{\textbf{Asset Price}} \\
            \cmidrule(lr){4-5} \cmidrule(lr){6-7}
             & (Interval) & & \textbf{Asset A} & \textbf{Asset B} & \textbf{$P_A$} & \textbf{$P_B$} \\
            \midrule
            
            \textbf{Initialization} & 1768873664 & Create Pool & 100 & 100 & 1.00 & 1.00 \\
            \arrayrulecolor{black!30}\midrule 
            
            \multirow{2}{*}{\textbf{Liquidity Mint}} & 1768873714 & \multirow{2}{*}{Mint(100A, 100B)} & \up{200} & \up{200} & \neu{1.00} & \neu{1.00} \\
             & (+50 blocks) & & {\scriptsize (+100\%)} & {\scriptsize (+100\%)} & {\scriptsize (-)} & {\scriptsize (-)} \\
            \arrayrulecolor{black!30}\midrule
            
            \multirow{2}{*}{\textbf{Token Swap}} & 1768873814 & \multirow{2}{*}{Swap(20B $\to$ A)} & \up{220} & \down{180} & \down{0.82} & \up{1.22} \\
             & (+100 blocks) & & {\scriptsize (+10\%)} & {\scriptsize (-10\%)} & {\scriptsize (-18\%)} & {\scriptsize (+22\%)} \\
            
            \arrayrulecolor{black}\bottomrule
        \end{tabular}%
    } 
    \vspace{-0.3cm}
\end{table}

Decentralized Finance (DeFi) \cite{auer2023technology} has evolved into a robust ecosystem, with Automated Market Makers (AMMs) serving as the foundational infrastructure. Unlike traditional limit order book models \cite{rocsu2009dynamic}, AMMs facilitate automated, peer-to-peer execution via smart contract-governed liquidity pools. This paradigm has driven innovation across decentralized exchanges (e.g., Uniswap V3 \cite{Uniswap2025}), lending protocols (e.g., Morpho \cite{morpho2026protocol}) and yield trading (e.g., Pendle \cite{Pendle2025}). By enabling continuous liquidity without centralized intermediaries, AMMs have significantly enhanced market efficiency.
One key feature of AMMs is their deterministic pricing mechanism, where the exchange rate is governed strictly by the reserve ratio of assets in the liquidity pool. While execution is driven by on-chain events, price fluctuations occur only when this ratio is altered. As shown in Table \ref{tab:amm_evolution}, during the \textit{Liquidity Mint} phase, reserves increase symmetrically (+100\%); however, since the asset proportion remains invariant, the price persists at $P_A=1.00$. Conversely, the \textit{Token Swap} event induces an asymmetric shift—increasing Asset A while burning Asset B—thereby modifying the reserve ratio. Consequently, the asset reprices to 0.82. This comparison validates that price volatility in AMMs is exclusively a derivative of ratio-altering events rather than aggregate liquidity changes.

Despite the importance of event-level analysis, there is still a lack of comprehensive event-level datasets for different on-chain protocols \cite{fan2022differential,xu2022met,muzammil2025poorest}. Furthermore, an in-depth analysis of on-chain transaction types remains largely underexplored. Bridging this gap offers critical insights for market traders. For example, by predicting the type of the next transaction, traders can deterministically infer the directional movement of the asset price within the liquidity pool, thereby identifying enhanced opportunities for profit and arbitrage.
To address data scarcity, we introduce a generalized pipeline designed for automated extraction and standardization of on-chain event data. To validate the robustness and universality of our framework, we curated a comprehensive dataset spanning a 21-month period from Jan 2024 to Sep 2025. We instantiated this pipeline across four AMMs protocols: Uniswap v3 (decentralized exchange), Aave and Morpho (lending), and Pendle (yield trading) to ensure coverage of diverse mechanisms and complex event logic.

\begin{figure*}
    \centering
    \includegraphics[width=.9\linewidth]{image/framework.png}
    \caption{Event-aware forecasting framework for AMM protocols. \textcolor{red}{(1)} Blockchain data pipeline extracts standardized on-chain events; \textcolor{red}{(2)} TPP model processes event sequences for joint prediction; \textcolor{red}{(3)} Uncertainty-weighted loss enhances temporal accuracy; \textcolor{red}{(4)} Downstream applications leverage predictions for yield trading strategies.}
    \label{fig:demo1}
\end{figure*}

{Additionally, } beyond dataset construction, we embrace the modeling challenge of forecasting both next \emph{transaction type} and \emph{occurrence block height} within this yield-trading environment. Conventional temporal point process (TPP) models have been extensively applied to continuous-time data \cite{daley2008introduction,Hawkes1971SpectraOS}, such as financial events \cite{hasbrouck1991measuring}, where timestamps follow wall-clock units and events are single-labeled. However, our setting departs from this norm in two critical ways. Firstly, time is measured in discrete block heights, which exhibit heavy-tailed gaps. Secondly, events are not atomic but may involve concurrent transaction types within the same block. These characteristics undermine standard intensity-based formulations and in calibration when ported to on-chain data. For stadnard neural TPP models, we found that while standard training achieves reasonable event type accuracy, it produces large block height errors, reflected in high RMSE, which severely limits usability in fast-moving DeFi markets where timing is critical.

To address this, we extend the TPP framework by augmenting the log-likelihood objective with an additional regression term on block height gaps, weighted by homoscedastic uncertainty. This design allows the model not only to fit the overall time distribution but also to minimize numerical deviations in prediction, a property crucial for downstream valuation of PT/YT lifecycles. As shown in our experiments, this modification substantially reduces timing error while maintaining event-type accuracy across TPP families.



\section{RELATED WORK}

\begin{table*}[htbp]
  \centering
  \caption{Comparisons with other blockchain datasets and ours.}
  \label{tab:comparisons}
  \begin{tabular}{@{}llcccc@{}}
    \toprule
    \textbf{Dataset} & \textbf{Field} & \textbf{AMM focus} & \textbf{Datasets detail} & \textbf{Granularity} & \textbf{Application richness} \\
    \midrule
    \cite{cernera2023token,li2024cryptotrade} & transaction & --- & on-chain transfers with off-chain social signals  & tx-level & price forecasting \\
    \cite{han2025pulsereddit} & transaction & --- & Reddit-driven high-frequency transaction & tx-level & high frequency trading \\
    \cite{shamsi2022chartalist,luo2024multi} & transaction & --- & cross-chain transaction graph & tx-level & social-driven trading \\
    \cite{hu2024zipzap,muzammil2025poorest} & fraud & --- & fraud detection & amount-level & defi risk \\
    \cite{xu2022met,long2024coinclip} & MEME COIN & --- & rug-pull txs and price manipulation & coin-level & meme coin risk \\
    \cite{zhou2024artemis,costa2023show} & NFT & --- & NFT minting, bidding, and airdrop hunting & NFT-level & NFT price and airdrop \\
    \cite{feng2025cryptomixer,ni2024money} & AMM & \checkmark & AMM pools transaction and profit  & tx-level & AMM profit max \\
    ours & \textcolor{red}{AMM} & \checkmark & AMM pools event driven & event-level & lifecycle modeling \\
    \bottomrule
  \end{tabular}
\end{table*}

\subsection{Blockchain Datasets}
A rich body of literature has introduced blockchain-specific datasets that span a variety of analytical tasks. Early efforts concentrated on the trading ecosystem \cite{chen2024finml,zhang2023live}: several works fuse on-chain transfers with off-chain social signals to forecast cryptocurrency prices \cite{cernera2023token,li2024cryptotrade}. Reddit-driven high-frequency discussions were later harvested to create a micro-structure-oriented dataset for crypto-asset trading \cite{han2025pulsereddit}. To support graph-based learning, a cross-chain transaction graph that covers Bitcoin, Ethereum, and BSC has also been released \cite{shamsi2022chartalist,luo2024multi}. Beyond price prediction, fraud detection \cite{hu2024zipzap,muzammil2025poorest} and account profiling \cite{wang2023ex,li2025multi} have received significant attention. More recently, fine-grained datasets targeting Meme Coins  have been proposed to detect rug-pull events and characterize price manipulation \cite{xu2022met,long2024coinclip}. In the NFT space, asset-level datasets that capture the full lifecycle—minting, bidding, and airdrop hunting—have been constructed to predict post-launch floor-price dynamics and hunter behavior \cite{zhou2024artemis,costa2023show}. In Table \ref{tab:comparisons}, we show the uniqueness of our datasets with a comparison with previous work.

Crucially, no publicly available dataset to date systematically labels or summarizes the events generated by on-chain yield-trading activities, leaving a conspicuous gap for data-driven research on fixed-income-style DeFi instruments.

\subsection{AI for Blockchain}
Transaction prediction on blockchains have long been central research themes. Classical financial indicators \cite{ibikunle2020more} and statistical models \cite{rzayev2019state} , however, fail to capture micro-level dynamics of individual users or single transactions in a decentralized environment. Motivated by the success of recurrent neutral networks \cite{sherstinsky2020fundamentals}, researchers have transplanted deep learning to on-chain data: convolutional networks extract short-term candlestick motifs \cite{sun2024graphsage}, while CryptoMixer \cite{feng2025cryptomixer} designs a two-stream MLP mixer to capure user trading behavior patterns. Yet these studies remain entrenched in a conventional-finance mindset, overlooking the blockchain-native coupling of liquidity pools, instantaneous trades, and price updates. Recently, many work focus on liquidity providers \cite{heimbach2022risks,fan2022differential,fan2023towards}. \cite{fan2021strategic} empirically measure historical LP payoffs and emit rebalancing strategies, whereas \cite{ni2024money} search for the globally optimal liquidity-mining allocation under a fixed set of tokens and pools. 

Notably, no existing work has yet employed models to forecast the next-transaction type, leaving the micro-temporal evolution of pool state an open prediction problem for data-driven DeFi research.

\section{DATASETS}
\label{data}
\subsection{Background}

\textbf{Automated Market Makers (AMMs).} Unlike traditional financial markets that rely on Limit Order Books (LOB) to match buyers and sellers, AMMs utilize liquidity pools to facilitate permissionless, automated digital asset trading. In this model, users (Liquidity Providers) deposit asset pairs into a smart contract, where prices are determined not by an order book, but by a deterministic mathematical formula. Formally, let a liquidity pool consist of two crypto-assets, $X$ and $Y$, with reserves denoted as $x$ and $y$, respectively. The trading mechanism is governed by a pricing function $f(\cdot)$ that maintains a constant invariant $k$. When a user executes a trade—swapping an amount $\delta_x$ for $\delta_y$—the protocol ensures the state transition satisfies:
$$f(x, y) = f(x+\delta_x, y-\delta_y) = k$$

To capture a comprehensive view of the DeFi ecosystem, we selected four representative protocols:

\textbf{Uniswap V3.} As the largest and most active Decentralized Exchange (DEX), it provides the foundational dataset for high-frequency spot trading events.

\textbf{Aave \& Morpho.} Representing the lending sector, these protocols are crucial for analyzing events related to liquidity supply and borrowing interest.

\textbf{Pendle.} The leading protocol in the yield trading domain. Its inclusion allows us to analyze complex pricing events derived from the separation of principal and yield tokens.

More details refer to the Appendix \ref{AMM Formulation}.

\subsection{Data Collection} To construct a high-fidelity dataset encompassing diverse DeFi sectors, we operate a \emph{read-only} Ethereum full node via the Chainstack infrastructure.\footnote{\url{https://ethereum-mainnet.core.chainstack.com/}} All data are collected exclusively through the public JSON-RPC interface, ensuring both verifiability and on-chain reproducibility of our dataset.

Event-Based Extraction Pipeline. Rather than processing the global ERC-20 \textsc{Transfer} stream, which is semantically ambiguous and computationally prohibitive due to its volume, we implement a targeted event-driven extraction strategy. We focus on \emph{application-level} events explicitly emitted by the smart contracts of the target protocols (Uniswap V3, Aave, Morpho, and Pendle) during state transitions. To ensure universality across different AMM and mechanisms, we categorize relevant events into three types:

\textbf{Trading Events:} Events representing the exchange of assets (e.g., \texttt{Swap} in Uniswap and Pendle). 

\textbf{Liquidity Events:} Events reflecting the modification of market depth (e.g., \texttt{Mint/Burn} in AMMs). 

\textbf{State Update Events:} Protocol-specific signals affecting pricing logic (e.g., \texttt{UpdateImpliedRate} in Pendle). 

More details refer to the Appendix \ref{Dataset Pipelines}.

\subsection{Data Structure} We organized the raw logs from Uniswap, Aave, Morpho, and Pendle into a standardized, human-readable JSON format. Each entry comprises a universal header for temporal alignment (e.g., timestamps, block numbers) and a protocol-specific payload that translates complex smart contract logic into interpretable metrics—such as trade volumes, lending interest rates, or yield adjustments. This design decouples financial analysis from underlying technical complexities, allowing researchers to treat on-chain events simply as high-granularity time-series data. Detailed schema definitions are provided in Appendix~\ref{sec:dataset_description}.

\subsection{Data Analysis}
\subsubsection{Dataset Overview} We introduce a high-fidelity, block-level DeFi dataset capturing 8,917,353 on-chain events from the Ethereum mainnet, spanning from Jan 1, 2024, to Sep 16, 2025 (block height: 18,908,896–23,374,292). The dataset aggregates micro-structural dynamics across four distinct protocol architectures—Uniswap V3, Aave, Pendle, and Morpho—covering a total of 359 liquidity pools (Pendle: 296, Aave: 53, Uniswap: 5, Morpho: 5) \footnote{The disparity in market counts reflects our rigorous selection of the most representative and active pools for each protocol to ensure high-fidelity analysis, rather than indicating the relative importance of the protocols themselves.}.

\begin{figure*}
    \centering
    \includegraphics[width=0.99\textwidth]{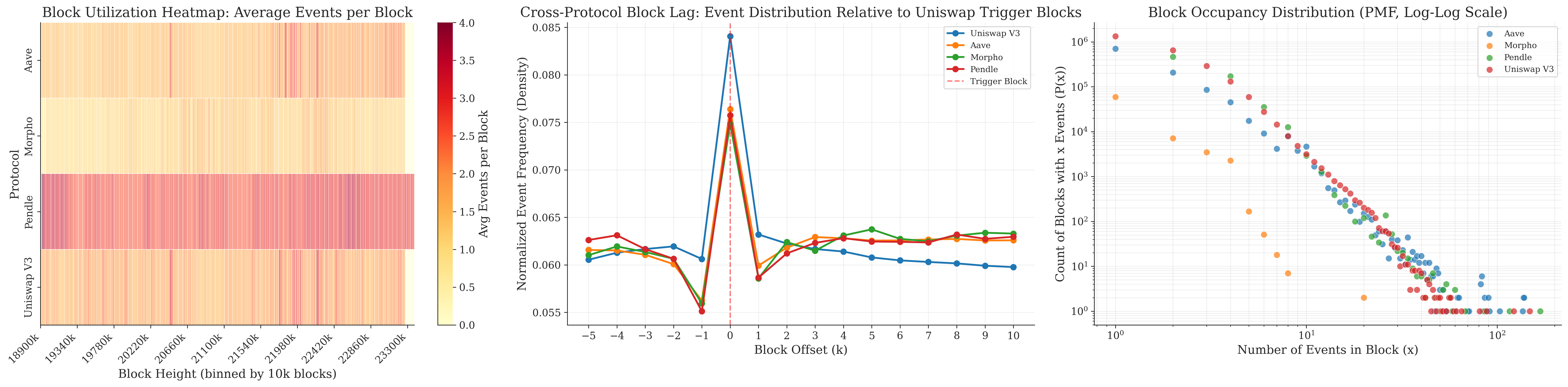}
    \caption{Block utilization, event synchronization, and occupancy distributions in DeFi protocols}
    \label{fig:block}
\end{figure*}

\subsubsection{Dataset Details} As shown in the Figure \ref{fig:block} a, the heatmap visualizes the temporal density of on-chain interactions by the metric Average Events per Block (EpB) across binned block heights (10k blocks/bin). The visualization reveals a stark dichotomy in protocol behavior: Pendle exhibits a high-intensity, continuous utilization pattern (consistent deep red), attributable to its frequent state-updating mechanism. Conversely, Uniswap V3 demonstrates "bursty" utilization characterized by intermittent vertical red stripes, indicating that its block space consumption is highly correlated with exogenous market volatility shocks. Aave and Morpho maintain a sparse, low-entropy state (predominantly yellow), reflecting the lower frequency of lending interactions compared to AMM trading or yield tokenization updates.

As shown in the Figure \ref{fig:block} b, this plot investigates the causal synchronization between protocols using Conditional Event Probability relative to a Uniswap V3 Trigger Block (defined as the 95th percentile of swap volume). The probability density function $P(Event_{protocol} | Block_{offset}=k)$ exhibits a sharp, synchronous peak at $k=0$ for all observed protocols (Aave, Morpho, Pendle). This indicates that liquidity events and interest rate updates occur atomically within the same block as the spot price shock. The rapid decay at $k=1$ suggests an extremely short information diffusion window. The absence of a "right-shifted" peak implies that off-chain keepers and MEV bots effectively bridge price disparities instantly, negating any significant inter-block latency.

In the Figure \ref{fig:block} c, the scatter plot presents the Probability Mass Function (PMF) of block occupancy on a Log-Log scale, confirming that transaction frequency follows a Power Law distribution ($P(x) \propto x^{-\alpha}$). Uniswap V3 (red) exhibits the heaviest tail, extending to $x > 200$, which signifies its distinct capability to absorb extreme "Black Swan" traffic surges. The horizontal alignment of data points at $y=1$ ($10^0$) in the tail region represents discrete, unique congestion events—blocks that historically occurred exactly once with such high transaction counts. In contrast, Morpho (orange) shows the steepest decay, indicating a more predictable and sparse usage pattern without the extreme outliers seen in the AMM layer.

\section{TASK FORMULATION}

The on-chain behavior of 4 protocol markets is inherently discrete and governed by operational time (i.e., block height and event index) rather than continuous wall-clock time. While physical time gaps between blocks may vary, the state of these AMMs is invariant to these durations and changes \emph{only} upon the execution of a valid transaction. Typically, one block height corresponds to 12 to 15 seconds. Consequently, modeling based on physical time introduces artificial sparsity and sampling bias, whereas treating the event sequence as the native temporal dimension aligns strictly with the Ethereum Virtual Machine (EVM) state transition logic. By adopting this event-driven perspective, our model filters out the stochastic noise of block intervals and focuses on the causal chain of liquidity interactions. This approach ensures that the modeled dynamics reflect the system’s true microstructure, where actionable signals and price updates occur exclusively at discrete block boundaries, regardless of the physical time elapsed.

Based on the data collected in Sec.~\ref{data}, we formalize our research problem as an event-driven modeling task. In Pendle markets, events happen solely when certain lifecycle events are induced, which are \textit{Mint}, \textit{Burn}, \textit{Swap}, and \textit{UpdateImpliedRate}. During intervals between these periods, contract states are constant, so event sequences, not smooth wall-clock time, are best to characterize the progression of the system. Measuring along block height guarantees proper order of executions and evades anomalies of wall-clock time. This definition thus encodes the real dynamic of market activity and offers a natural foundation for arbitrage.

\paragraph{Event Sequence and Modeling Objective.}
We model the evolution of each pool as a \emph{marked event sequence} along the block height axis.  
For an asset $a$, the transaction history can be denoted as 
\[
\mathcal{S}^a = \{(t_i, y_i)\}_{i=1}^{L_a},
\]
where $t_i \in \mathbb{N}$ denotes the block height of the $i$-th event, and $y_i \in \mathcal{Y}$ is its (atomic or composite) mark (event type).
The event history up to block $t$ is 
\[
\mathcal{H}_t = \{(t_j, y_j): t_j < t\},
\]
and the inter-block gap is $\Delta_i = t_i - t_{i-1} \in \mathbb{N}^+$.

Our benchmark task is to predict the \textbf{next mark} $y_{i+1}$ and the \textbf{next event time} $t_{i+1}=t_i+\Delta_{i+1}$ given $\mathcal{H}_{t_i}$.
We adopt the \emph{Marked Temporal Point Process (TPP)} framework~\cite{Hawkes1971SpectraOS,mei2017neural}, in which the joint distribution of times and marks is modeled conditionally on historical observations.
Note that representation for $\Delta_{i+1}$ and $t_{i+1}$ is isomorphic in this framework. This formulation provides a probabilistic foundation for forecasting on-chain behavior and understanding the microstructure dynamics of Pendle AMM markets.


\section{BENCHMARK}
Pendle pool states change only when lifecycle events occur. Between events, reserves and implied rates are constant. The history is therefore a sequence of block indexed event times and marks $(t_i, y_i)$. And TPPs fit this setting. They model the conditional law of the next gap and mark given $\mathcal{H}_{t_i}$, capturing dependence across events. This yields a principled likelihood for joint time–type prediction and a clear basis for benchmarking on block height.

Hence, this section will concisely illustrate the objective of TPP and its standard likelihood loss for training. Besides, due to the numerical characteristics of block height, directly adopting standard loss would amplify the time prediction flaws inherent in TPP architecture. Therefore, this section will also introduce the loss function improvements we implemented, resulting in the new loss function, Uncertainty Weighted MSE (UWM). 

\subsection{Temporal Point Process}
\label{subsec:joint}
Drawing on \cite{Hawkes1971SpectraOS,mei2017neural}
, we get the following standard TPP objective. Conditioned on $\mathcal{H}_{t_i}$, we factor the joint of the next event as

\[
\begin{aligned}
p(t_{i+1}, y_{i+1} \mid \mathcal{H}_{t_i})
&= p\big(y_{i+1}\mid \mathcal{H}_{t_i}\big)\cdot p\big(t_{i+1}\mid \mathcal{H}_{t_i}\big) \\
&= \pi_i[y_{i+1}] \cdot f_{\psi}({\Delta}_{i+1}\mid \mathcal{H}_{t_i}).
\end{aligned}
\]

where the $\pi_i[y_{i+1}]$ is a function that represents the distribution over the next event type. The $f_{\psi}({\Delta}_{i+1}\mid \mathcal{H}_{t_i})$ is the conditional density of the next gap for times, involving cumulative hazard and survival relationship. The sequence log-likelihood over $[t_0,t_{L_a}]$ is the usual sum of \emph{event} terms and \emph{time} terms. In the hazard formulation, the negative log-likelihood for times reduces to $-\sum_i \log f_{\psi}({\Delta}_{i+1}\mid \mathcal{H}_{t_i})$. This is equivalent in value to the standard intensity-based objective, and easier to implement with basis hazards under block time. Hence, the standard loss can be written as $Loss = -\sum_i \log \pi_i[y_{i+1}] -\sum_i \log f_{\psi}(\Delta_{i+1}\mid \mathcal{H}_{t_i})$.

\subsection{Uncertainty Weighted MSE Loss}
\label{subsec:objective}
The blockchain market is characterized by rapid trading, meaning that any disruptive transaction event requires investors to respond promptly. In this context, the occurrence of such events, reflected in block height, is particularly crucial.  

However, pure standard likelihood loss can fit distributions well but still yield large numeric errors on block gaps when tails are heavy. It is hypothesized that this is due to TPP placing excessive emphasis on the disparities in distributions, while concomitantly disregarding the discrepancies in numerical residuals. We therefore add MSE to balance and combine it as:

\begin{enumerate}
\item Mark likelihood $\ell_{\text{mark}}=-\sum_i \log \pi_i[y_{i+1}]$,
\item Time likelihood $\ell_{\text{time}}=-\sum_i \log f_{\psi}(\Delta_{i+1}\mid \mathcal{H}_{t_i})$,
\item Time regression $\ell_{\text{mse}}=\sum_i \big(\widehat{{\Delta}}_{i+1}-{\Delta}_{i+1}\big)^2$.
\end{enumerate}

We aggregate them with \emph{Homoscedastic Uncertainty weighting} \cite{kendall2018multi}. Let $s=(s_1,s_2,s_3)$ be unconstrained scalars and define $\sigma_m=\mathrm{softplus}(s_m)>0$. The objective for this Uncertainty Weighted MSE (UWM) loss is
\[
\label{eq:uwm}
\begin{aligned}
Loss
&= \frac{1}{2}\sum_{m=1}^{3}\Big(\sigma_m^{-2}\,\ell_m + \log \sigma_m^2\Big) \\
\ell_1 &= \ell_{\text{mark}}, \ell_2 = \ell_{\text{time}},\ell_3 = \ell_{\text{mse}}.
\end{aligned}
\]
This scheme learns the relative weights from data and stabilizes training across assets with the large block gaps. The idea is orthogonal to the encoder and compatible with RNN-, attention-, or ODE-style TPPs. Likelihood-based training for TPPs and joint time or mark modeling are well established. Therefore, we adopt 8 representative TPP models with UWM loss for our task.  

\section{EXPERIMENT}
\label{sec:experiments}

In this section, we evaluate the performance of our proposed loss on the task of next-event prediction over yield-bearing blockchain assets, with the objective of jointly \textit{mark} (event type) multi-classification and \textit{event time} (in block height units) forecasting. We describe the processed dataset, baseline models, training settings, and metrics, followed by detailed comparisons, ablation studies. 

\subsection{Experimental Setup}
\noindent\textbf{Dataset.} From the dataset we build, each transaction involves the transaction type and the block height of occurrence. We choose two representative protocols \footnote{Through exploratory data analysis, it was determined that Uniswap V3, Aave, and Morpho are all high-density event data. Hence, Uniswap V3 was selected as the representative dataset.} as the dataset for our task, high density event dataset (Uniswap V3) and sparse event dataset (Pendle). Also due to the highly imbalanced sequence length for each asset in both protocols, we choose to split the long sequence into multiple short sequences. After processing and chronological ordering, the Pendle dataset comprises 295 distinct assets and 1,624 sequences of transaction series data. These are partitioned into training, validation, and testing splits of 1,141, 239, and 244 sequences, respectively. Across all splits, the dataset contains 484{,}004 individual transactions (340{,}178 training, 71{,}040 validation, 72{,}786 testing).
And for Uniswap V3, it comprises 10,596 transaction series data, involving 7,417, 1,589 and 1,590 sequences for training, validation and testing. Across all splits, it contains 2{,}632{,}486 individual transactions (1{,}840{,}860 training, 393{,}956 validation, 397{,}670 testing)



\noindent\textbf{Temporal structure.} We treat block height as the temporal axis. Hence, the Pendle dataset contains 483{,}994 inter-event intervals with a mean interval of 48.5 blocks, a median of 16 blocks, and a long-tailed distribution with a maximum gap of 1,426 blocks. This heavy-tailed and sparse pattern confirms that some assets experience bursty transaction phases separated by long idle spans, making prediction more challenging than representative neural TPP benchmarks. Conversely, the Uniswap V3 dataset involves 2,632,486 inter-event intervals with a mean interval of 2.6 blocks and a median of 1.0 block, exhibiting a dense pattern. Consequently, block heights (8-digit numbers) were not normalized due to the presence of large event intervals within the temporal structure.

\noindent\textbf{Event types.} The processed dataset finally involves 5 transaction types from raw dataset, indicating Swap in, Swap out, Mint, Burn and UpdateImpliedRate. However, concurrency is common, with multiple types occurring simultaneously. To capture this phenomenon, we represent each unique co-occurrence as a single categorical label, yielding 31 distinct event-type combinations. This multi-label collapse reflects the true stochasticity of blockchain execution environments while remaining tractable for sequence modeling.

\noindent\textbf{Evaluation splits.} Following standard practice, all metrics are reported on the held-out test set. Validation is used for hyperparameter tuning and early stopping. Splits are performed chronologically at the sequence level to avoid event-level leakage. We additionally verify that no assets overlap across splits.


\subsection{Baselines and Metrics}
\noindent\textbf{Baseline models.} We benchmark against eight state-of-the-art temporal point process (TPP) models spanning recurrent, attention, and neural-ODE families:

\begin{itemize}
    \item \textbf{Neural Hawkes Process (NHP)} \cite{mei2017neural}: Uses a continuous time LSTM whose hidden state evolves between events to produce neurally self-modulating intensities. 
    \item \textbf{RMTPP} \cite{rmtpp}: Encodes event history with an RNN and models the conditional intensity as a learned nonlinear function of that history to jointly predict next event.
    \item \textbf{Self-Attentive Hawkes Process (SAHP)} \cite{sahp}: Links self attention to Hawkes modeling and introduces time shifted positional encodings that convert inter-event intervals into sinusoidal phase shifts for history dependent intensities. 
    \item \textbf{Transformer Hawkes Process (THP)} \cite{thp}: Replaces recurrence with multi-head self-attention and continuous-time temporal encodings to summarize long histories.
    \item \textbf{Attentive NHP (AttNHP)} \cite{attnhp}: Attends over past events with continuous time positional encodings and, uniquely, embeds each possible event $e@t$ so that $\lambda_e(t)$ is derived from the candidate event’s own embedding.
    \item \textbf{IntensityFree} \cite{intensityfree}: Dispenses with intensity parameterization by learning the conditional inter-event-time density via normalizing flows and a log-normal mixture that allows closed form sampling.  
    \item \textbf{FullyNN} \cite{fullynn}: Parameterizes the integrated conditional intensity with a feed-forward network conditioned on history and obtains $\lambda^*(t)$ by differentiation.
    \item \textbf{ODETPP}: Combines Neural ODEs with continuous-time normalizing flows \cite{ode} without spatial elements to model high-fidelity temporal mark distributions with likelihoods.
\end{itemize}

\noindent\textbf{Training objective.} All baselines are trained with the standard negative log-likelihood (NLL), decomposed into event-type likelihood and inter-event time likelihood. For our contribution, we extend the objective by adding a mean-squared error (MSE) term on time prediction, balanced by learnable uncertainty weights $\sigma_{i}$.

\noindent\textbf{Metrics.} Following Temporal Point Process Benchmark convention \cite{xue2023easytpp}, we adapt the following evaluation metrics:
\begin{itemize}
    \item \textbf{Type Accuracy}: classification accuracy over 31 event-types.
    \item \textbf{Time RMSE}: root mean squared error between predicted and observed inter-event times. The smaller root mean squared error can ensure the occurrence time of next indicated event transaction without delay.
    \item \textbf{OTD:} (Optimal Transport Distance) \cite{mei2019imputing}: evaluation metric for long-horizon prediction. A sequence-level edit distance capturing both transaction type and occurrence time mismatches, evaluated with horizon $H=5$.
\end{itemize}

\subsection{Standard TPP Loss}
The \textbf{Baseline (NLL)} column in Table~\ref{tab:combined_uwm_results} report baseline performance under the conventional NLL objective on the testing set for both Pendle and Uniswap V3. Across datasets, \emph{IntensityFree} remains a strong non-intensity baseline due to its closed-form likelihood evaluation, while intensity-based baselines rely on Monte Carlo integration and can incur approximation errors. In particular, IntensityFree achieves the lowest time RMSE on Pendle (81.32), and it is tied for the best RMSE on Uniswap V3 (4.64, matched by SAHP).
\noindent Under NLL, Pendle is substantially more challenging for several intensity-based architectures, with RMSE exploding for RMTPP (769.14), THP (958.37), FullyNN (665.56), and ODETPP (6815.47), reflecting temporal miscalibration under blockchain-style bursts.

\begin{table*}[h]
\centering
\small
\setlength{\tabcolsep}{8pt} 
\renewcommand{\arraystretch}{1.1} 
\caption{Time RMSE ($\downarrow$) comparison on Pendle and Uniswap V3 Datasets. We compare baseline models against their Uncertainty Weighted MSE (UWM) variants. Values are mean $\pm$ std over 5 random seeds. 
\textcolor{red}{Red} indicates improvement, 
\textcolor{green!60!black}{green} indicates degradation, 
and \textbf{\textcolor{red}{bold red}} marks dramatic improvement (> 30\%).}
\label{tab:combined_uwm_results}
\begin{tabular}{lcccc}
\toprule
\multirow{2}{*}{\textbf{Model}} & \multicolumn{2}{c}{\textbf{Pendle Dataset}} & \multicolumn{2}{c}{\textbf{Uniswap V3 Dataset}} \\
\cmidrule(lr){2-3} \cmidrule(lr){4-5}
& Baseline (NLL) & With UWM Loss & Baseline (NLL) & With UWM Loss \\
\midrule
NHP 
& 108.89 $\pm$ 1.84 & \textcolor{red}{94.28 $\pm$ 0.21} 
& 5.26 $\pm$ 0.24 & \textcolor{red}{4.70 $\pm$ 0} \\
RMTPP 
& 769.14 $\pm$ 147.25 & \textbf{\textcolor{red}{95.55 $\pm$ 0.1}} 
& 5.28 $\pm$ 0.07 & \textcolor{green!60!black}{6.59 $\pm$ 0.54} \\
SAHP 
& 93.87 $\pm$ 3.85 & \textcolor{green!60!black}{93.98 $\pm$ 0.29} 
& 4.64 $\pm$ 0.05 & \textcolor{red}{4.63 $\pm$ 0.02} \\
THP 
& 958.37 $\pm$ 62.97 & \textbf{\textcolor{red}{94.50 $\pm$ 1.28}} 
& 5.06 $\pm$ 0.03 & \textcolor{red}{5.06 $\pm$ 0} \\
AttNHP 
& 100.63 $\pm$ 2.22 & \textcolor{red}{96.73 $\pm$ 3.32} 
& 25.49 $\pm$ 1.13 & \textbf{\textcolor{red}{5.03 $\pm$ 0.33}} \\
IntensityFree 
& 81.32 $\pm$ 2.50 & --- 
& 4.64 $\pm$ 0.16 & --- \\
FullyNN 
& 665.56 $\pm$ 100.64 & \textbf{\textcolor{red}{91.08 $\pm$ 1.71}} 
& 108.72 $\pm$ 1.61 & \textcolor{red}{105.43 $\pm$ 0.29} \\
ODETPP 
& 6815.47 $\pm$ 1266.82 & \textbf{\textcolor{red}{277.07 $\pm$ 293.57}} 
& 5.20 $\pm$ 0.08 & \textcolor{green!60!black}{7.47 $\pm$ 2.39} \\
\bottomrule
\end{tabular}
\end{table*}

\noindent In summary, these results establish a challenging NLL baseline across two distinct regimes. Pendle exposes large temporal errors for multiple intensity-based models, whereas Uniswap V3 is generally easier but still admits sharp failures for specific architectures. his motivates our introduction of an uncertainty weighted MSE loss term to better balance likelihood fitting and time regression under bursty dynamics. The complete results for Type accuracy and OTD ($H=5$) is shown in table \ref{tab:abl_baselin_uwm_result} in Appendix C.

\subsection{Uncertainty Weighted MSE Loss}
The ``With UWM Loss'' column in Table~\ref{tab:combined_uwm_results} presents performance when augmenting the objective with our proposed uncertainty weighted MSE. On Pendle, temporal accuracy improves substantially across nearly all models. NHP’s RMSE decreases from 109.48 to 94.33, and RMTPP achieves its best balance at 95.591 RMSE with OTD reduction and higher type accuracy. Even THP, previously unstable, improves dramatically from 1053.819 RMSE to 94.58. The RMSE metric saw an average improvement of \textbf{56.41\%}. This suggests that the auxiliary regression term stabilizes temporal predictions across architectures. Besides, based on the conversion between block height and Greenwich Mean Time (GMT), an RMSE of 93 can be approximated as the result of a 19 minute floating error at the time of occurrence. On Uniswap V3, where several baselines already cluster at low RMSE (e.g., SAHP 4.64), UWM yields smaller changes (SAHP 4.64 $\rightarrow$ 4.63; NHP 5.26 $\rightarrow$ 4.70), but can still provide a strong corrective effect when the baseline collapses. Most notably, AttNHP drops from 25.49 to 5.03 (an $\sim$80\% improvement).

\begin{figure*}[t]
  \centering
  \begin{subfigure}{0.33\textwidth}
    \includegraphics[width=\linewidth]{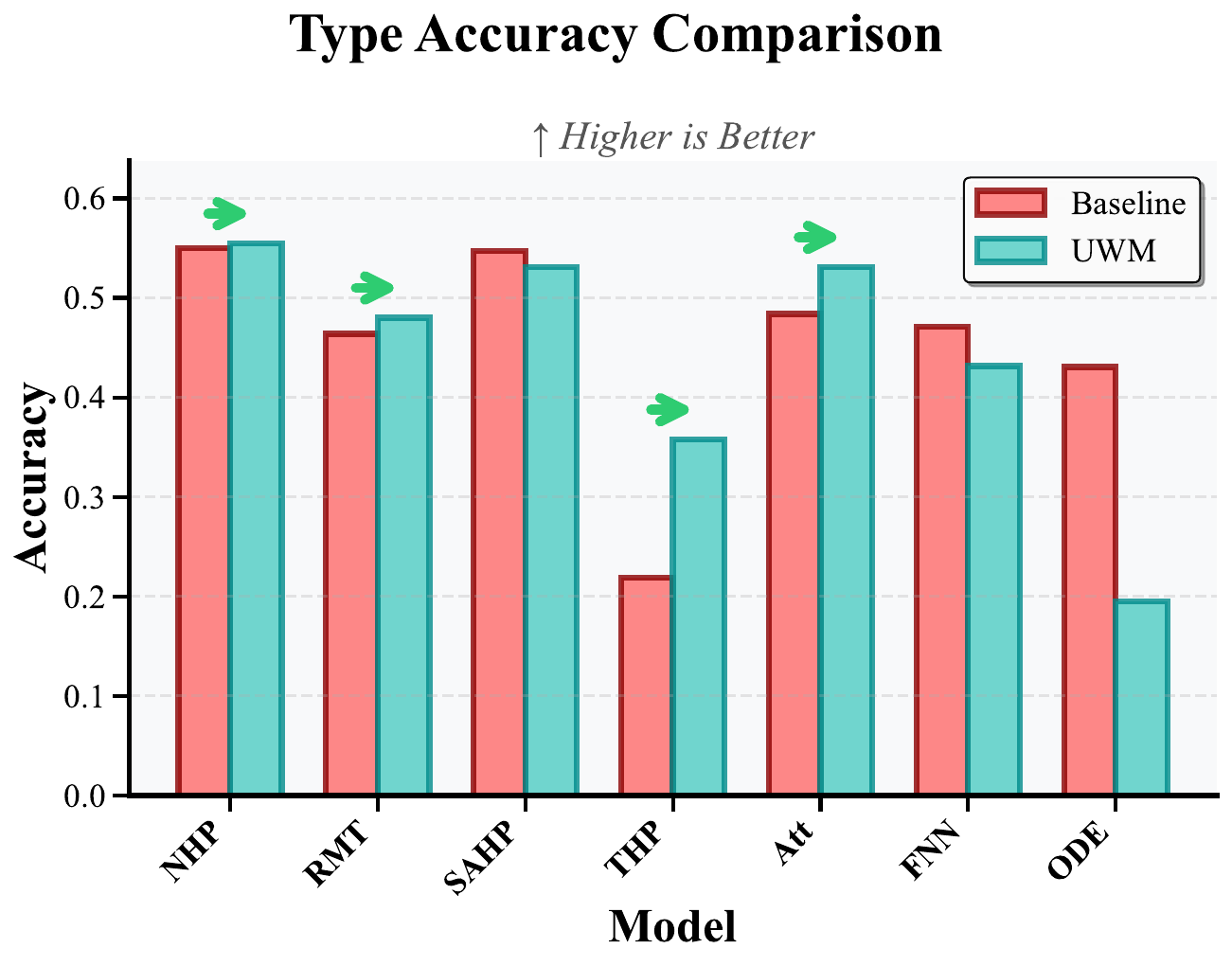}%
  \end{subfigure}\hfill
  \begin{subfigure}{0.33\textwidth}
    \includegraphics[width=\linewidth]{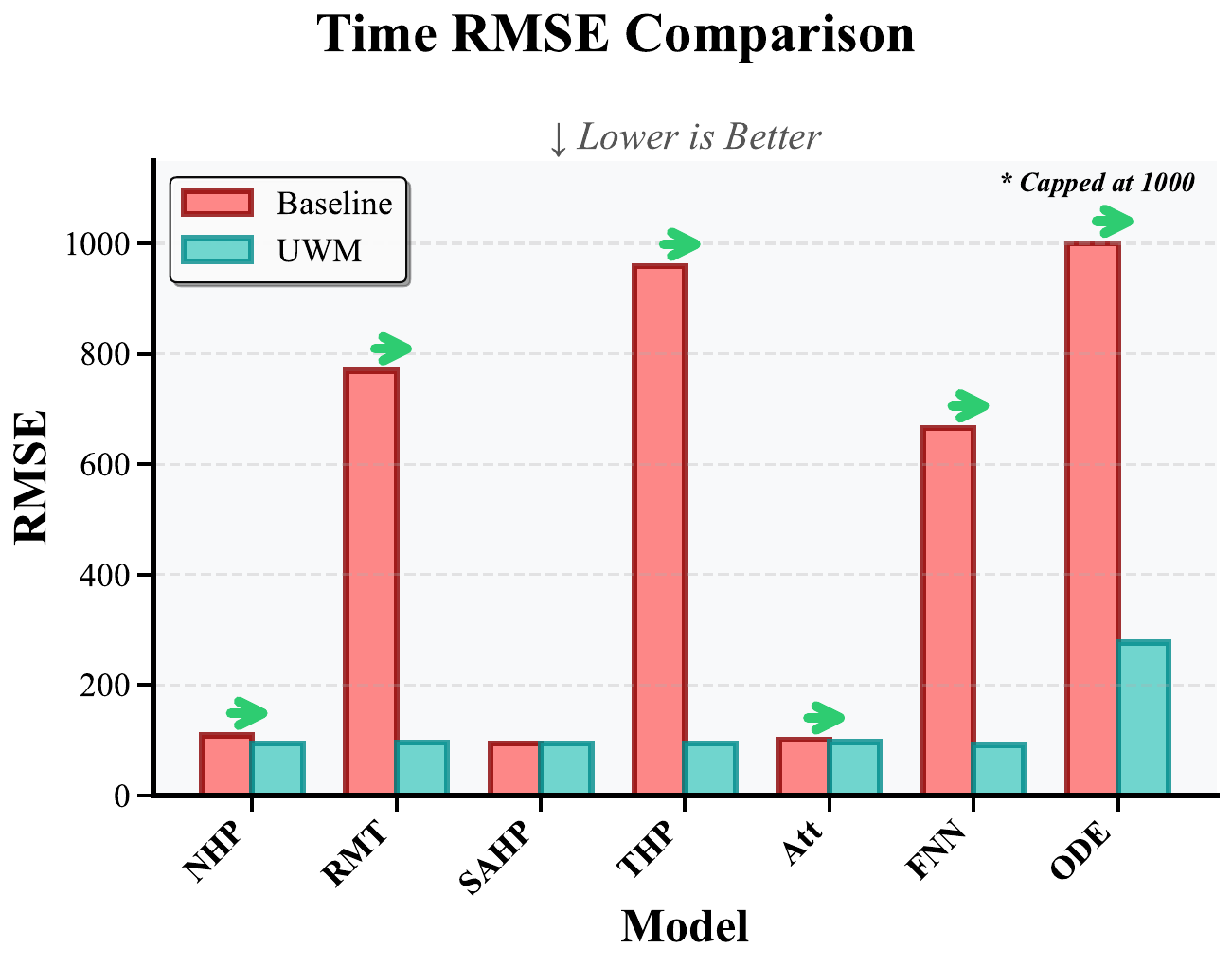}%
  \end{subfigure}\hfill
  \begin{subfigure}{0.33\textwidth}
    \includegraphics[width=\linewidth]{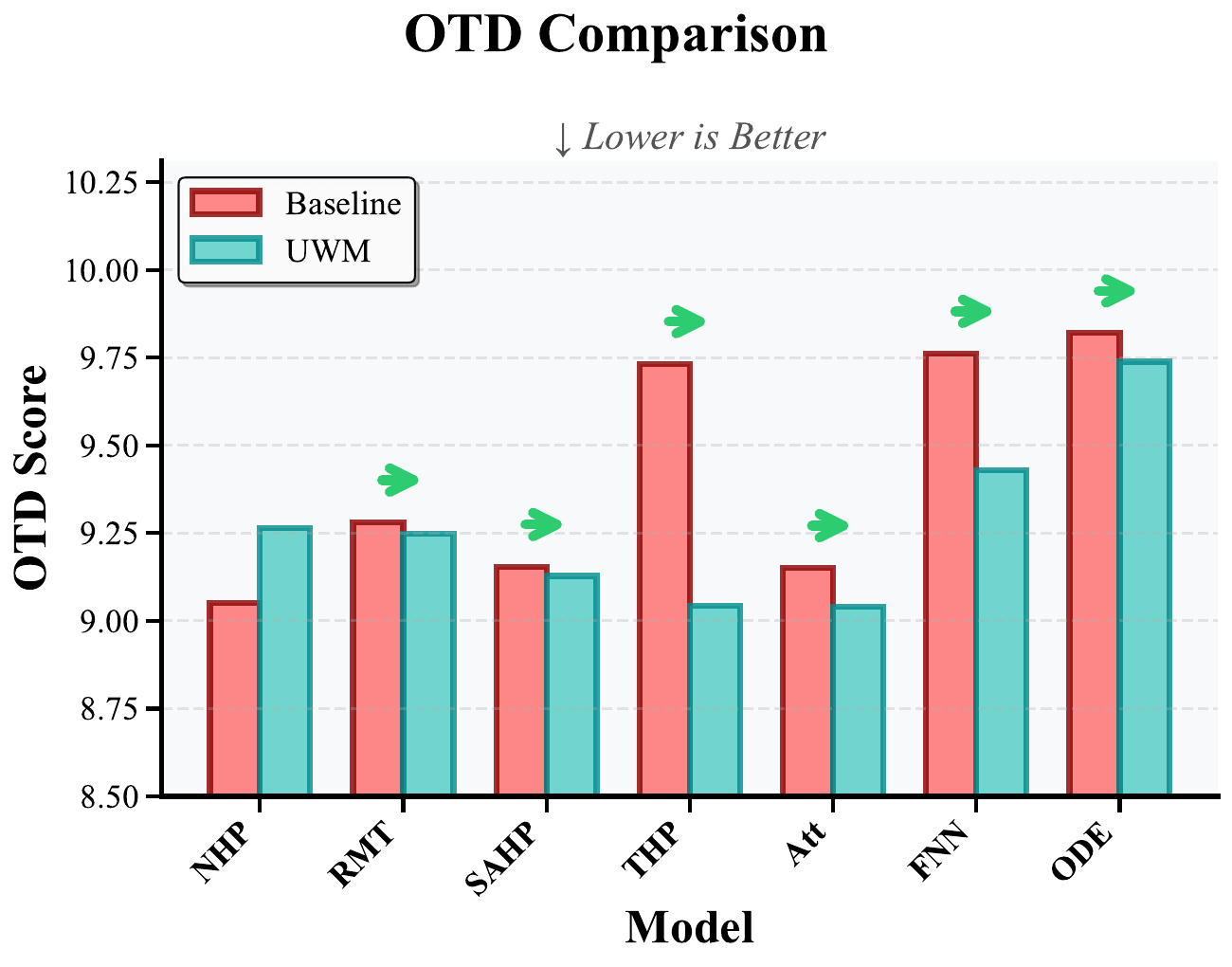}%
  \end{subfigure}
  \caption{Improvement from Metrics.}%
  \label{fig:improvment}
\end{figure*}

\noindent From Fig \ref{fig:improvment}, the green arrow indicates that the model with UWM outperforms the baseline in Pendle dataset. These improvements validate our claim that distributional likelihood alone is insufficient for block-level time forecasting, especially in sparse event datasets. The uncertainty weighted MSE complements the probabilistic fit by directly minimizing numeric deviations. The IntensityFree model, which doesn't use intensity, has a similar implicit function for penalizing the residual between block height. This means it is not included in the UWM loss.


\subsection{Ablation Studies}
We next investigate the contribution of each loss component, the effect of learned uncertainty, and the sensitivity to horizon length.

\begin{figure*}[t]
  \centering
  \begin{subfigure}{0.33\textwidth}
    \includegraphics[width=\linewidth]{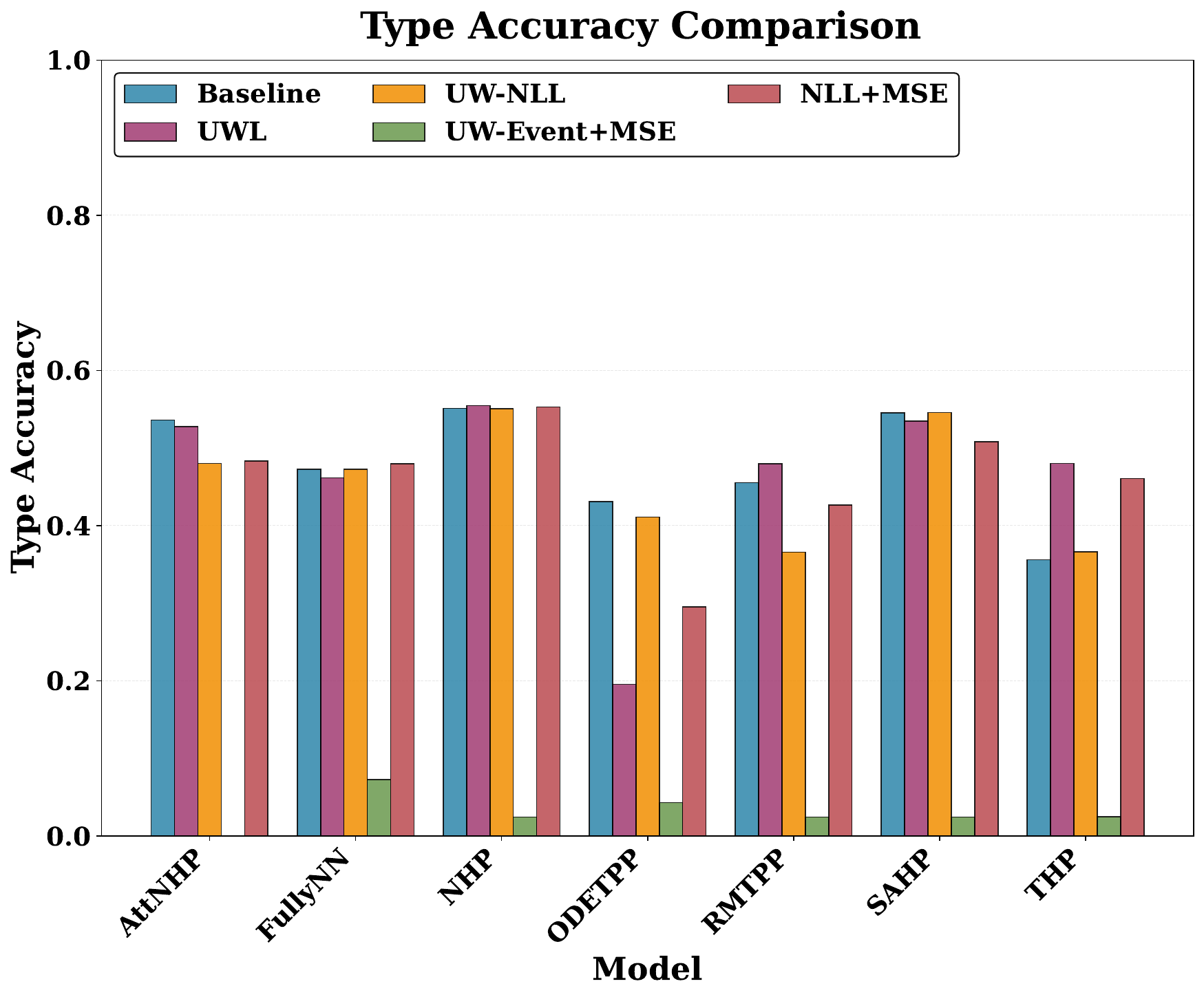}%
    \label{fig:three-wide:a}%
  \end{subfigure}\hfill
  \begin{subfigure}{0.33\textwidth}
    \includegraphics[width=\linewidth]{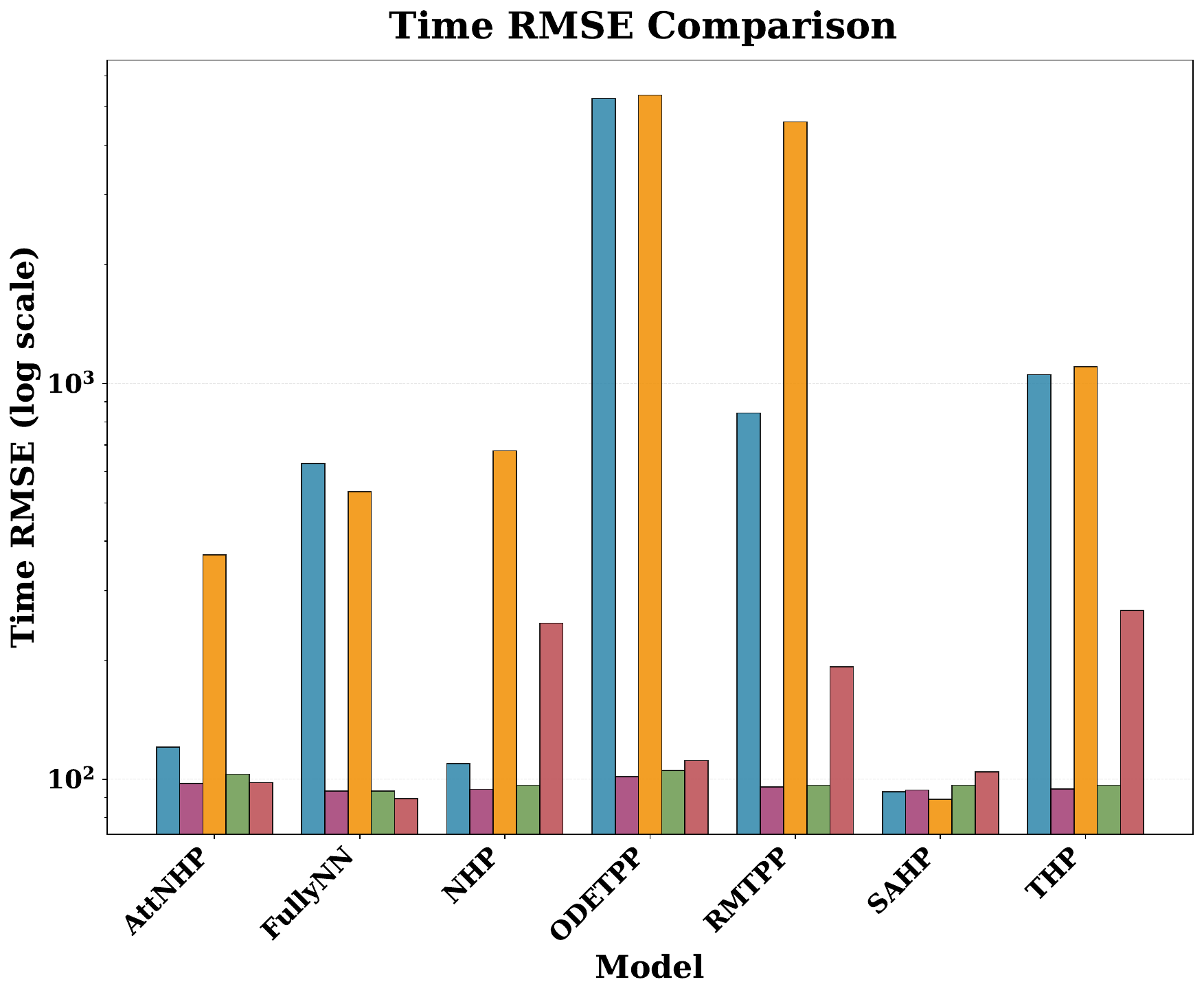}%
    \label{fig:three-wide:b}%
  \end{subfigure}\hfill
  \begin{subfigure}{0.33\textwidth}
    \includegraphics[width=\linewidth]{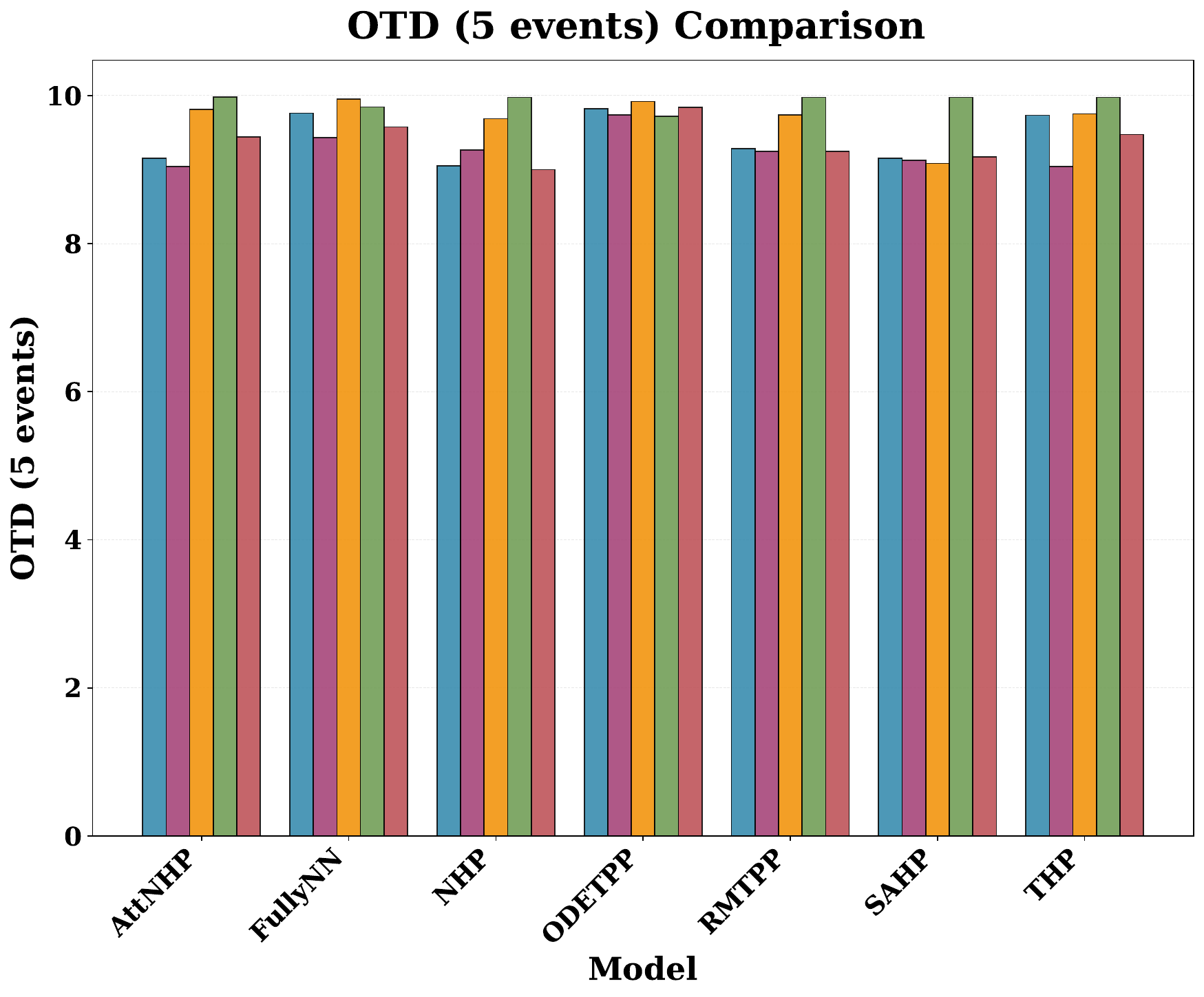}%
    \label{fig:three-wide:c}%
  \end{subfigure}
  \caption{Ablation study: Effect of MSE term and Uncertainty Weighting.}%
  \label{fig:Ablationstudy}
\end{figure*}


\noindent\textbf{Effect of the MSE term.} Define "UW-nll" as the UWM loss without $\ell_{\text{mse}}$, and "UW-Event+MSE" as the UWM loss without $\ell_{\text{time}}$. Shown in Figure \ref{fig:Ablationstudy}, removing the MSE term leads to degradation in temporal accuracy, confirming its necessity. Conversely, removing $\ell_{\text{time}}$ while retaining MSE yields under-calibrated likelihoods, though with competitive RMSE. A balanced combination is essential.


\noindent\textbf{Uncertainty learning.} 
Shown in Figure \ref{fig:Ablationstudy}, fixing $\sigma$ values rather than learning them degrades performance, particularly when time variance is high. Learned uncertainties adaptively rescale contributions, yielding more stable training across assets.



\begin{figure}[!htbp]
    \centering
    \includegraphics[width=0.9\linewidth]{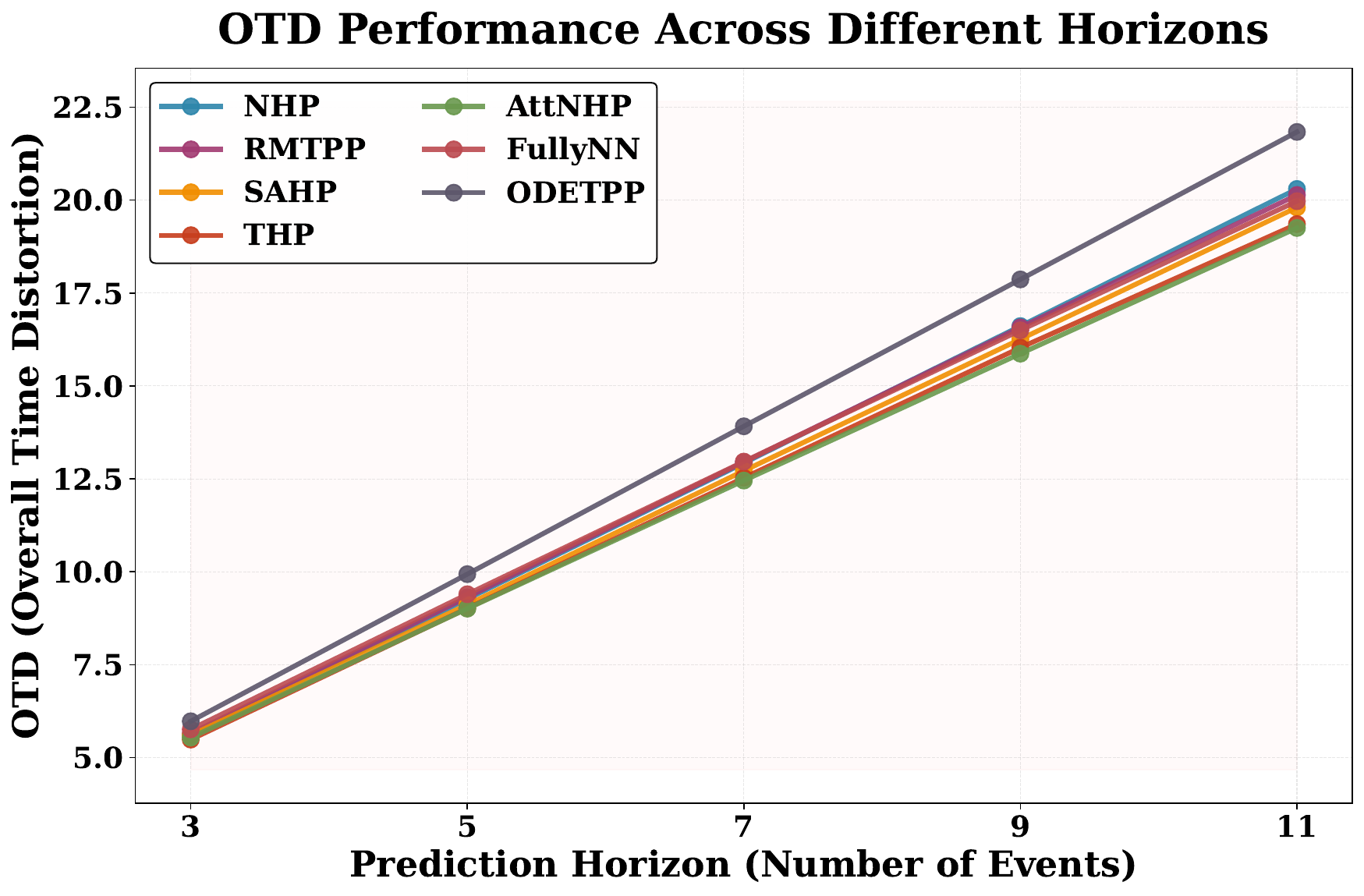}
    \caption{Ablation study: Horizon sensitivity.}
    \label{fig:otdablation}
\end{figure}

\noindent\textbf{Horizon sensitivity.} Figure \ref{fig:otdablation} shows the OTD metrics across horizons $H=\{3,3,7,11\}$ with UWM series model. This type of model represents a linear and stable growth from short to long horizon predictions, building a scaling law between OTD and prediction horizon across different TPPs. 

\section{DOWNSTREAM APPLICATIONS}
\subsection{Applications}
This section explores potential downstream applications of the \textbf{Pendle-Event} dataset.

\textbf{Yield Trading.} Modeling \textbf{Mint}, \textbf{Burn}, \textbf{Swap}, and \textbf{UpdateImpliedRate} sequences as temporal processes enables predicting short-term PT/SY ratio fluctuations, which directly reflect Pendle's implied yields. For instance, anticipating \textbf{Swap} events that decrease the PT/SY ratio signals lower future yields, prompting traders to increase exposure to Yield Tokens (YT). Conversely, a rising ratio suggests higher future yield value, guiding optimal allocation between PT and YT positions. This event-driven framework provides a data-driven foundation for adaptive yield speculation and portfolio rebalancing based on predicted market transitions.

\textbf{Arbitrage.} Structured event data facilitates identifying arbitrage opportunities by forecasting event types and magnitudes, such as traded volume or implied rate shifts. By simulating anticipated price adjustments, agents can detect predictable deviations between PT, YT, and underlying SY tokens. This predictive layer allows for real-time arbitrage execution, capturing temporary mispricings before market corrections occur. 

\subsection{Case Study: Event-Aware Trading in sUSDe}

In this section, we present a case by Pendle sUSDe pools\cite{pendle2025susdepool}. 

Consider a Pendle AMM pool for sUSDe (Ethena) that initially holds 1,000 PT and 2,000 SY (ratio = 0.50). A predicted \textbf{Swap} in 2 blocks will add 200 PT and remove 400 SY, yielding 1,200 PT and 1,600 SY (ratio = 0.75). In Pendle's design, a higher PT/SY ratio means more PT is available relative to SY, so each PT is worth less SY on the market. Equivalently, the price of PT in terms of SY falls. Because PT price is inversely related to its embedded fixed yield, this shift implies a higher implied yield on holding PT to maturity. (Pendle's implied APY rises when PT is cheap: when the Implied APY is high, PT is cheaper (relative to the underlying token), and the resulting fixed yield APY is higher.) In this example, PT goes from 2 SY (initial price = $2000\,\text{SY} / 1000\,\text{PT}$) to about 1.333 SY (final price = $1600/1200$), a $\sim$33\% drop. This larger discount means the fixed payout at maturity (1 PT = 1 SY of underlying) represents a higher yield. 

\textbf{Yield Trading Strategy.} Given the higher implied yield, a yield trader might rebalance their position. For example, one could sell PT and increase exposure to YT. Selling PT realizes capital at the now-lower price, and buying YT captures the rising future yield stream. This is a classic yield-speculation move in Pendle: investors buy YT when they expect the asset's yield to climb. Concretely, suppose a trader holds 100 PT from before. After the predicted ratio change, each PT is worth fewer SY, so selling those 100 PT yields 133.3 SY (since $100 \times 1.333$).  Such a reallocation is supported by Pendle's design: YT holders profit if the underlying APY exceeds market expectations, while PT locks in a fixed rate.

\textbf{Arbitrage Opportunity: Buying PT and Selling into the Pool.} The predicted price shift also creates a short-term arbitrage opportunity between the Pendle pool and external markets. Suppose that on an external exchange or another liquidity pool, PT trades at 1.50 SY per PT. Before the swap event, the Pendle pool priced PT at 2 SY, so one could buy PT externally (1.50 SY) and sell into the pool (2 SY) for profit. For example:
\begin{itemize}
    \item Step 1: Buy 100 PT off-platform for 150 SY (100 * 1.50).
    \item Step 2: Immediately sell these 100 PT into the Pendle pool at the pool price of 2 SY, receiving 200 SY.
    \item Profit: $200 - 150 = 50$ SY (minus fees).
\end{itemize}

After the swap (when pool PT price falls to 1.333 SY), the arbitrage flips direction. Now the pool is cheaper: one could buy PT \textit{in the pool} at 1.333 SY and sell externally at 1.50 SY. For instance:
\begin{itemize}
    \item Step 1: Use SY to buy 100 PT from the pool at 1.333 SY each (cost 133.3 SY).
    \item Step 2: Sell those 100 PT on the external market at 1.50 SY, receiving 150 SY.
    \item Profit: $150 - 133.3 = 16.7$ SY (minus fees).
\end{itemize}

Either way, the price discrepancy is arbitraged away. In Pendle's mechanism, such trades leverage the fact that PT is normally \textit{discounted} relative to the underlying (reflecting its fixed yield). By buying low and selling high across markets (or vice versa), arbitrageurs enforce price efficiency. Pendle's pools therefore allow traders to exploit exactly these discrepancies: investors can utilize the price differences between different maturities of PT/YT for arbitrage.

\subsection{Auxiliary Information}  
Due to space limitations, we provide comprehensive supplementary materials in the appendices. \textbf{Appendix \ref{AMM}} details the mathematical formulation of Automated Market Makers. \textbf{Appendix \ref{sec:dataset_description}} elaborates our dataset construction methodology, presenting protocol-specific extraction pipelines, event schema definitions, and in-depth statistical analyses of temporal patterns across 4 protocols. \textbf{Appendix \ref{results}} reports complete evaluation results for all eight TPP baselines under both standard NLL and our proposed UWM loss across multiple random seeds. \textbf{Appendix \ref{sec:appendixC}} documents the full experimental setup, including hardware specifications, hyperparameters, and reproducibility protocols. 

\section{CONCLUSION}
Despite the central role of Automated Market Makers (AMMs) in Decentralized Finance (DeFi) for driving asset pricing through deterministic reserve mechanisms, existing research has largely overlooked the micro-structural event dynamics and lacked fine-grained modeling frameworks. To address these gaps, we present a comprehensive dataset comprising 8.9 million events from four representative protocols—Pendle, Uniswap V3, Aave, and Morpho—covering the period from January 2024 to September 2025, with precise annotations of transaction types and block heights. Furthermore, we propose the Uncertainty Weighted MSE (UWM) Loss function, which adaptively integrates block-interval regression into the Temporal Point Process (TPP) objective. Empirical results across eight advanced TPP architectures demonstrate that our approach reduces time prediction error by an average of 56.41\% while maintaining high event-type accuracy. Our work establishes a critical benchmark for event-aware forecasting and provides a foundation for future research in multi-protocol synchronization and the optimization of high-frequency trading strategies in on-chain ecosystems.

\newpage
\bibliographystyle{unsrt}   
\bibliography{sample-base}

@String{Computer = "{IEEE} Computer" }

@String{Springer = "Springer-Verlag" }

@misc{Uniswap2025,
  title        = {Uniswap},
  url          = {https://app.uniswap.org/},
  note         = {Accessed: 2025-09-25}
}

@misc{Pendle2025,
  title        = {Pendle},
  url          = {https://www.pendle.finance/},
  note         = {Accessed: 2025-09-25}
}

@article{ibikunle2020more,
  title={More heat than light: Investor attention and bitcoin price discovery},
  author={Ibikunle, Gbenga and McGroarty, Frank and Rzayev, Khaladdin},
  journal={International Review of Financial Analysis},
  volume={69},
  pages={101459},
  year={2020},
  publisher={Elsevier}
}

@article{rzayev2019state,
  title={A state-space modeling of the information content of trading volume},
  author={Rzayev, Khaladdin and Ibikunle, Gbenga},
  journal={Journal of Financial Markets},
  volume={46},
  pages={100507},
  year={2019},
  publisher={Elsevier}
}

@article{sun2024graphsage,
  title={GraphSAGE with deep reinforcement learning for financial portfolio optimization},
  author={Sun, Qiguo and Wei, Xueying and Yang, Xibei},
  journal={Expert Systems with Applications},
  volume={238},
  pages={122027},
  year={2024},
  publisher={Elsevier}
}

@inproceedings{feng2025cryptomixer,
  title={CryptoMixer: Fine-grained market information-aware MLP Networks for Individual Cryptocurrency Trading Prediction},
  author={Feng, Tingsheng and Shen, Zhihao and Zhao, Xi and Lu, Xiaoni and Zhou, Yuyang},
  booktitle={Proceedings of the 31st ACM SIGKDD Conference on Knowledge Discovery and Data Mining V. 2},
  pages={603--614},
  year={2025}
}

@inproceedings{ni2024money,
  title={Money Never Sleeps: Maximizing Liquidity Mining Yields in Decentralized Finance},
  author={Ni, Wangze and Yiwei, Zhao and Sun, Weijie and Chen, Lei and Cheng, Peng and Zhang, Chen Jason and Lin, Xuemin},
  booktitle={Proceedings of the 30th ACM SIGKDD Conference on Knowledge Discovery and Data Mining},
  pages={2248--2259},
  year={2024}
}

@inproceedings{fan2022differential,
  title={Differential Liquidity Provision in Uniswap v3 and Implications for Contract Design},
  author={Fan, Zhou and Marmolejo-Coss{\'\i}o, Francisco J and Altschuler, Ben and Sun, He and Wang, Xintong and Parkes, David},
  booktitle={Proceedings of the Third ACM International Conference on AI in Finance},
  pages={9--17},
  year={2022}
}

@article{xue2023easytpp,
  title={Easytpp: Towards open benchmarking temporal point processes},
  author={Xue, Siqiao and Shi, Xiaoming and Chu, Zhixuan and Wang, Yan and Hao, Hongyan and Zhou, Fan and Jiang, Caigao and Pan, Chen and Zhang, James Y and Wen, Qingsong and others},
  journal={arXiv preprint arXiv:2307.08097},
  year={2023}
}

@inproceedings{cernera2023token,
  title={Token spammers, rug pulls, and sniper bots: An analysis of the ecosystem of tokens in ethereum and in the binance smart chain ($\{$$\{$$\{$$\{$$\{$BNB$\}$$\}$$\}$$\}$$\}$)},
  author={Cernera, Federico and La Morgia, Massimo and Mei, Alessandro and Sassi, Francesco},
  booktitle={32nd USENIX Security Symposium (USENIX Security 23)},
  pages={3349--3366},
  year={2023}
}

@inproceedings{li2024cryptotrade,
  title={CryptoTrade: A reflective LLM-based agent to guide zero-shot cryptocurrency trading},
  author={Li, Yuan and Luo, Bingqiao and Wang, Qian and Chen, Nuo and Liu, Xu and He, Bingsheng},
  booktitle={Proceedings of the 2024 Conference on Empirical Methods in Natural Language Processing},
  pages={1094--1106},
  year={2024}
}

@article{han2025pulsereddit,
  title={PulseReddit: A Novel Reddit Dataset for Benchmarking MAS in High-Frequency Cryptocurrency Trading},
  author={Han, Qiuhan and Wang, Qian and Yoshikawa, Atsushi and Yamamura, Masayuki},
  journal={arXiv preprint arXiv:2506.03861},
  year={2025}
}

@article{shamsi2022chartalist,
  title={Chartalist: Labeled graph datasets for utxo and account-based blockchains},
  author={Shamsi, Kiarash and Victor, Friedhelm and Kantarcioglu, Murat and Gel, Yulia and Akcora, Cuneyt G},
  journal={Advances in Neural Information Processing Systems},
  volume={35},
  pages={34926--34939},
  year={2022}
}

@article{luo2024multi,
  title={Multi-Chain Graphs of Graphs: A New Approach to Analyzing Blockchain Datasets},
  author={Luo, Bingqiao and Zhang, Zhen and Wang, Qian and He, Bingsheng},
  journal={Advances in Neural Information Processing Systems},
  volume={37},
  pages={28490--28514},
  year={2024}
}

@article{chen2024finml,
  title={Finml-chain: A blockchain-integrated dataset for enhanced financial machine learning},
  author={Chen, Jingfeng and Deng, Wanlin and Chen, Dangxing and Zhang, Luyao},
  journal={arXiv preprint arXiv:2411.16277},
  year={2024}
}

@article{zhang2023live,
  title={Live graph lab: Towards open, dynamic and real transaction graphs with nft},
  author={Zhang, Zhen and Luo, Bingqiao and Lu, Shengliang and He, Bingsheng},
  journal={Advances in Neural Information Processing Systems},
  volume={36},
  pages={18769--18793},
  year={2023}
}

@inproceedings{hu2024zipzap,
  title={Zipzap: Efficient training of language models for large-scale fraud detection on blockchain},
  author={Hu, Sihao and Huang, Tiansheng and Chow, Ka-Ho and Wei, Wenqi and Wu, Yanzhao and Liu, Ling},
  booktitle={Proceedings of the ACM Web Conference 2024},
  pages={2807--2816},
  year={2024}
}

@inproceedings{muzammil2025poorest,
  title={The poorest man in babylon: A longitudinal study of cryptocurrency investment scams},
  author={Muzammil, Muhammad and Pitumpe, Abisheka and Li, Xigao and Rahmati, Amir and Nikiforakis, Nick},
  booktitle={Proceedings of the ACM on Web Conference 2025},
  pages={1034--1045},
  year={2025}
}

@article{wang2023ex,
  title={Ex-graph: A pioneering dataset bridging ethereum and x},
  author={Wang, Qian and Zhang, Zhen and Liu, Zemin and Lu, Shengliang and Luo, Bingqiao and He, Bingsheng},
  journal={arXiv preprint arXiv:2310.01015},
  year={2023}
}

@article{li2025multi,
  title={Multi-source Multi-level Multi-token Ethereum Dataset and Benchmark Platform},
  author={Li, Haoyuan and Zhang, Mengxiao and Li, Maoyuan and Li, Jianzheng and Yang, Junyi and Deng, Shuangyan and Zhang, Zijian and Liu, Jiamou},
  journal={arXiv preprint arXiv:2501.11906},
  year={2025}
}

@inproceedings{xu2022met,
  title={Met-meme: A multimodal meme dataset rich in metaphors},
  author={Xu, Bo and Li, Tingting and Zheng, Junzhe and Naseriparsa, Mehdi and Zhao, Zhehuan and Lin, Hongfei and Xia, Feng},
  booktitle={Proceedings of the 45th international ACM SIGIR conference on research and development in information retrieval},
  pages={2887--2899},
  year={2022}
}

@article{long2024coinclip,
  title={CoinCLIP: A Multimodal Framework for Assessing Viability in Web3 Memecoins},
  author={Long, Hou-Wan and Li, Hongyang and Cai, Wei},
  journal={arXiv preprint arXiv:2412.07591},
  year={2024}
}

@inproceedings{zhou2024artemis,
  title={Artemis: Detecting airdrop hunters in nft markets with a graph learning system},
  author={Zhou, Chenyu and Chen, Hongzhou and Wu, Hao and Zhang, Junyu and Cai, Wei},
  booktitle={Proceedings of the ACM Web Conference 2024},
  pages={1824--1834},
  year={2024}
}

@inproceedings{costa2023show,
  title={Show me your NFT and I tell you how it will perform: Multimodal representation learning for NFT selling price prediction},
  author={Costa, Davide and La Cava, Lucio and Tagarelli, Andrea},
  booktitle={Proceedings of the ACM Web Conference 2023},
  pages={1875--1885},
  year={2023}
}

@article{sherstinsky2020fundamentals,
  title={Fundamentals of recurrent neural network (RNN) and long short-term memory (LSTM) network},
  author={Sherstinsky, Alex},
  journal={Physica D: Nonlinear Phenomena},
  volume={404},
  pages={132306},
  year={2020},
  publisher={Elsevier}
}

@inproceedings{heimbach2022risks,
  title={Risks and returns of uniswap v3 liquidity providers},
  author={Heimbach, Lioba and Schertenleib, Eric and Wattenhofer, Roger},
  booktitle={Proceedings of the 4th ACM Conference on Advances in Financial Technologies},
  pages={89--101},
  year={2022}
}

@article{fan2021strategic,
  title={Strategic liquidity provision in uniswap v3},
  author={Fan, Zhou and Marmolejo-Cossio, Francisco and Moroz, Daniel J and Neuder, Michael and Rao, Rithvik and Parkes, David C},
  journal={arXiv preprint arXiv:2106.12033},
  year={2021}
}

@article{fan2023towards,
  title={Towards understanding governance tokens in liquidity mining: a case study of decentralized exchanges},
  author={Fan, Sizheng and Min, Tian and Wu, Xiao and Wei, Cai},
  journal={World Wide Web},
  volume={26},
  number={3},
  pages={1181--1200},
  year={2023},
  publisher={Springer}
}

@book{daley2008introduction,
  title={An introduction to the theory of point processes: volume II: general theory and structure},
  author={Daley, Daryl J and Vere-Jones, David},
  year={2008},
  publisher={Springer}
}

@article{Hawkes1971SpectraOS,
  title={Spectra of some self-exciting and mutually exciting point processes},
  author={Alan G. Hawkes},
  journal={Biometrika},
  year={1971},
  volume={58},
  pages={83-90},
  url={https://api.semanticscholar.org/CorpusID:14122089}
}

@article{hasbrouck1991measuring,
  title={Measuring the information content of stock trades},
  author={Hasbrouck, Joel},
  journal={The Journal of Finance},
  volume={46},
  number={1},
  pages={179--207},
  year={1991},
  publisher={Wiley Online Library}
}

@article{mei2017neural,
  title={The neural hawkes process: A neurally self-modulating multivariate point process},
  author={Mei, Hongyuan and Eisner, Jason M},
  journal={Advances in neural information processing systems},
  volume={30},
  year={2017}
}

@inproceedings{kendall2018multi,
  title={Multi-task learning using uncertainty to weigh losses for scene geometry and semantics},
  author={Kendall, Alex and Gal, Yarin and Cipolla, Roberto},
  booktitle={Proceedings of the IEEE conference on computer vision and pattern recognition},
  pages={7482--7491},
  year={2018}
}

@inproceedings{rmtpp,
  title={Recurrent marked temporal point processes: Embedding event history to vector},
  author={Du, Nan and Dai, Hanjun and Trivedi, Rakshit and Upadhyay, Utkarsh and Gomez-Rodriguez, Manuel and Song, Le},
  booktitle={Proceedings of the 22nd ACM SIGKDD international conference on knowledge discovery and data mining},
  pages={1555--1564},
  year={2016}
}

@inproceedings{sahp,
  title={Self-attentive Hawkes process},
  author={Zhang, Qiang and Lipani, Aldo and Kirnap, Omer and Yilmaz, Emine},
  booktitle={International conference on machine learning},
  pages={11183--11193},
  year={2020},
  organization={PMLR}
}

@inproceedings{thp,
  title={Transformer hawkes process},
  author={Zuo, Simiao and Jiang, Haoming and Li, Zichong and Zhao, Tuo and Zha, Hongyuan},
  booktitle={International conference on machine learning},
  pages={11692--11702},
  year={2020},
  organization={PMLR}
}

@article{attnhp,
  title={Transformer embeddings of irregularly spaced events and their participants},
  author={Yang, Chenghao and Mei, Hongyuan and Eisner, Jason},
  journal={arXiv preprint arXiv:2201.00044},
  year={2021}
}

@article{intensityfree,
  title={Intensity-free learning of temporal point processes},
  author={Shchur, Oleksandr and Bilo{\v{s}}, Marin and G{\"u}nnemann, Stephan},
  journal={arXiv preprint arXiv:1909.12127},
  year={2019}
}

@article{fullynn,
  title={Fully neural network based model for general temporal point processes},
  author={Omi, Takahiro and Aihara, Kazuyuki and others},
  journal={Advances in neural information processing systems},
  volume={32},
  year={2019}
}

@article{ode,
  title={Neural spatio-temporal point processes},
  author={Chen, Ricky TQ and Amos, Brandon and Nickel, Maximilian},
  journal={arXiv preprint arXiv:2011.04583},
  year={2020}
}

@article{rocsu2009dynamic,
  title={A dynamic model of the limit order book},
  author={Ro{\c{s}}u, Ioanid},
  journal={The Review of Financial Studies},
  volume={22},
  number={11},
  pages={4601--4641},
  year={2009}
}

@inproceedings{mei2019imputing,
  title={Imputing missing events in continuous-time event streams},
  author={Mei, Hongyuan and Qin, Guanghui and Eisner, Jason},
  booktitle={International Conference on Machine Learning},
  pages={4475--4485},
  year={2019},
  organization={PMLR}
}

@misc{pendle2025susdepool,
  author       = {{Pendle Finance}},
  title        = {sUSDe Pool --- Trade \& Zap Interface},
  howpublished = {\url{https://app.pendle.finance/trade/pools/0xe06c3b972ba630ccf3392cecdbe070690b4e6b55/zap/in?chain=plasma}},
  note         = {Accessed: 2025-10-07},
  year         = {2025}
}

@techreport{auer2023technology,
  title        = {The Technology of Decentralized Finance (DeFi)},
  author       = {Auer, Raphael and Haslhofer, Bernhard and Kitzler, Stefan and Saggese, Pietro and Victor, Friedhelm},
  institution  = {Bank for International Settlements},
  year         = {2023},
  month        = {January},
  number       = {1066},
  type         = {BIS Working Papers},
  url          = {https://www.bis.org/publ/work1066.pdf}
}

@misc{morpho2026protocol,
  author = {{Morpho Labs}},
  title  = {{Morpho}: Open Lending Infrastructure},
  year   = {2026},
  url    = {https://morpho.org/},
  note   = {Accessed: February 5, 2026}
}

\clearpage    
\onecolumn    
\appendix     


\newpage

\section{Auto Market Makers}
\label{AMM}
\subsection{AMM Formulation}
\label{AMM Formulation}
Unlike the Limit Order Book (LOB) mechanism prevalent in centralized exchanges, Decentralized Exchanges (DEXs) predominantly utilize the Automated Market Maker (AMM) protocol. An AMM acts as an algorithmic agent that provides continuous, 24/7 liquidity without the need for a specific counterparty. It replaces traditional "maker" orders with a deterministic pricing formula governed by a conservation function stored in a smart contract.

Protocol Participants: The AMM ecosystem primarily consists of three roles:

\begin{itemize}

\item \textbf{Liquidity Providers (LPs):} Agents who deposit assets (e.g., pairs of tokens) into a liquidity pool to facilitate trading, earning protocol fees (typically 0.3\%) in return.
\item \textbf{Traders:} Users who execute swaps with the liquidity pool.
\item \textbf{Arbitrageurs:} Sophisticated agents who monitor external market prices (CEXs) and execute trades to align the AMM price with the market price, thereby ensuring price efficiency.
\end{itemize}

Conservation Function: The generic class of these protocols is termed Constant Function Market Makers (CFMMs). A liquidity pool holding reserves of two assets, $x$ and $y$, is governed by a convex function $\psi(x, y)$ such that the value remains constant across a trade:

\begin{equation}\psi(x, y) = \psi(x - \Delta x, y + \Delta y) = k\end{equation}

where $\Delta x$ is the output amount withdrawn by the trader and $\Delta y$ is the input amount deposited. 

Pricing Mechanism:The spot price $P_x$ (the price of asset $X$ in terms of $Y$) is defined as the marginal rate of substitution at the current state, derived via the implicit function theorem:\begin{equation}\frac{\partial \psi}{\partial x} dx + \frac{\partial \psi}{\partial y} dy = 0 \implies P_x = - \frac{dy}{dx} = \frac{\partial_x \psi}{\partial_y \psi}\end{equation}For the standard Constant Product Market Maker (CPMM) where $xy=k$, the spot price is simply the ratio of reserves: $P_x = y/x$.

\subsection{AMM Insights} 

While the formulation provides the mechanism, the economic implications of AMMs differ significantly from traditional markets.Slippage and Execution Price:In AMMs, large orders traverse a significant portion of the bonding curve, resulting in \textit{slippage}—the difference between the spot price at the time of the order and the effective execution price.\begin{example}Consider a pool with reserves $x=100$ ETH, $y=40,000$ USDC ($k=4 \times 10^6$). The initial spot price is $P_{spot} = 400$ USDC/ETH. A trader purchasing $\Delta x = 10$ ETH effectively shifts the pool to $x'=90$. Based on $(x-\Delta x)(y+\Delta y)=k$, the new $y'$ requires $\Delta y \approx 4,444.44$ USDC.The effective price is $\bar{P} = 444.44$ USDC/ETH. The slippage is $\bar{P} - P_{spot} = 44.44$ USDC/ETH ($11.11\%$).\end{example}While positive slippage (favorable execution) is possible via mean-reversion, negative slippage is inherent to convex invariant curves.

Impermanent Loss (IL) vs. Arbitrage Loss (AL):LPs face specific risks distinct from holding assets.\begin{itemize}\item \textbf{Impermanent Loss (IL):} The opportunity cost of providing liquidity versus holding the assets outside the pool (HODL). If the external price shifts from $P_0$ to $P_1$, the pool rebalances. If prices revert to $P_0$, IL vanishes.\item \textbf{Arbitrage Loss (AL):} Also known as Loss-Versus-Rebalancing (LVR). This represents the absolute profit extracted by arbitrageurs. When the pool price lags behind the external market, arbitrageurs trade against the pool at a stale price.\end{itemize}

Event-Driven Price Dynamics: A critical and often overlooked distinction in DeFi forecasting is the fundamental nature of price updates. In traditional Centralized Finance (CeFi), market makers and high-frequency trading algorithms actively adjust quotes in milliseconds based on continuous off-chain information flows (news, sentiment, macro data). In stark contrast, AMMs operate as \textit{passive, deterministic state machines}. This passive nature dictates that the on-chain price is not a continuous function of time, but a discrete function of transaction events.

\begin{itemize}

    \item \textbf{Typology of Event Impact:} 
    Not all events impact the system identically. We distinguish between two primary classes of updates:
    \begin{itemize}
        \item \textit{Swap Events (First-Order Impact):} These events directly traverse the bonding curve $\psi(x,y)=k$, altering the ratio $y/x$ and thus causing an immediate shift in the spot price $P$. The magnitude of the shift is deterministic and dependent on the swap size relative to the pool depth.
        \item \textit{Liquidity Events (Second-Order Impact):} Mint and Burn events typically do not alter the spot price directly (assuming assets are deposited at the current ratio), but they alter the curvature of the bonding curve (changing $k$). This modifies the pool's \textit{price sensitivity} (or depth). A massive \texttt{Burn} event, for instance, makes the pool shallow, causing subsequent \texttt{Swap} events to trigger significantly larger price slippage.
    \end{itemize}

    \item \textbf{Arbitrage as the Information Bridge:} 
    Since the AMM is isolated from the real world, the "Event" serves as the sole bridge for information transmission. When off-chain prices shift, an arbitrage opportunity opens. The arrival of an arbitrage \texttt{Swap} event on-chain is the \textit{actualization} of that off-chain information. Therefore, forecasting the arrival and magnitude of these arbitrage events is functionally equivalent to forecasting the convergence of the on-chain price to the fair market value.
\end{itemize}

\newpage

\section{Dataset}
\label{sec:dataset_description}

\subsection{Dataset Notions}

We provide detailed descriptions of the event schemas and field definitions for each protocol in Tables \ref{tab:uniswap_events}, \ref{tab:aave_events}, \ref{tab:morpho_events}, and \ref{tab:pendle_events}.

\subsection{Dataset Pipelines}
\label{Dataset Pipelines}
Our data acquisition infrastructure leverages a read-only Ethereum full node via Chainstack's JSON-RPC interface to extract historical events from Aave V3, Morpho Blue, Uniswap V3, and Pendle spanning Jan 2024 to Sep 2025. Block-level temporal resolution is achieved by mapping timestamps to block heights through binary search on eth\_getBlockByNumber, while parallel processing via a 6–8 worker pool accelerates large-range scans. Transient RPC failures are mitigated through exponential backoff retries, ensuring fault-tolerant execution over multi-month ranges. For Aave, Morpho, and Uniswap V3, we employ event-log extraction using eth\_getLogs with ABI-derived topic filters (e.g., \textbf{Swap}, \textbf{Supply}, \textbf{Liquidate}), segmenting large ranges into 2,000–10,000 block chunks to avoid timeouts; raw logs are parsed and normalized by token decimals. Pendle requires a state-based approach: we invoke view functions (e.g., readState) via eth\_call every 100 blocks to snapshot market state and derive Principal Token (PT) and Yield Token (YT) prices according to Pendle's pricing model. All outputs are standardized into a unified JSON schema with block-aligned timestamps and decimal-normalized values, organized by asset/pool identifiers for efficient downstream analysis.

\subsection{Dataset Description}
This study utilizes high-fidelity blockchain event data extracted from the smart contracts of four major DeFi protocols: Uniswap V3, Aave, Morpho, and Pendle. The data covers a diverse range of liquidity pools and lending markets. Each dataset is structured in JSON format, chronologically ordered by block number. Common attributes across all datasets include \texttt{block\_number}, \texttt{timestamp}, \texttt{transaction\_hash}, and \texttt{log\_index}, ensuring precise temporal alignment and event ordering.


\subsubsection{Uniswap V3 Event Data}
\label{app:uniswap-data}

The Uniswap V3 dataset captures liquidity and trading activities within specific $\mathit{Token}_0$--$\mathit{Token}_1$ liquidity pools on Ethereum mainnet. We consider five major pools: \textbf{USDC-ETH}, \textbf{WBTC-ETH}, \textbf{WBTC-USDC}, \textbf{ETH-USDT}, and \textbf{WBTC-USDT}. The dataset comprises \textbf{4{,}861{,}101} events in total, spanning blocks approximately 18.9M--23.3M (circa January 2024--August 2025). The most active pool, USDC-ETH, contributes roughly 80\% of all events (3.9M); the remaining pools each contain between 137k and 507k events. This volume and coverage make the dataset representative of both high- and moderate-liquidity AMM venues.

\begin{itemize}
    \item \textbf{Event Types:}
    \begin{itemize}
        \item \textbf{Mint/Burn:} Liquidity provision and removal events. These alter the pool's active liquidity distribution and, indirectly, execution prices.
        \item \textbf{Swap:} Token exchange events. Each swap corresponds to a price-moving trade.
    \end{itemize}
    \item \textbf{Schema:} Key fields include \texttt{type}, \texttt{blockNumber}, \texttt{txHash}, and \texttt{timestamp}. It also tracks token flows via \texttt{amount0} and \texttt{amount1} (where negative values indicate outgoing assets and positive values indicate incoming ones).
\end{itemize}

\paragraph{Event composition.}

\begin{wrapfigure}{r}{0.5\textwidth} 
    \centering
    \vspace{-1.em}
    
    \includegraphics[width=\linewidth]{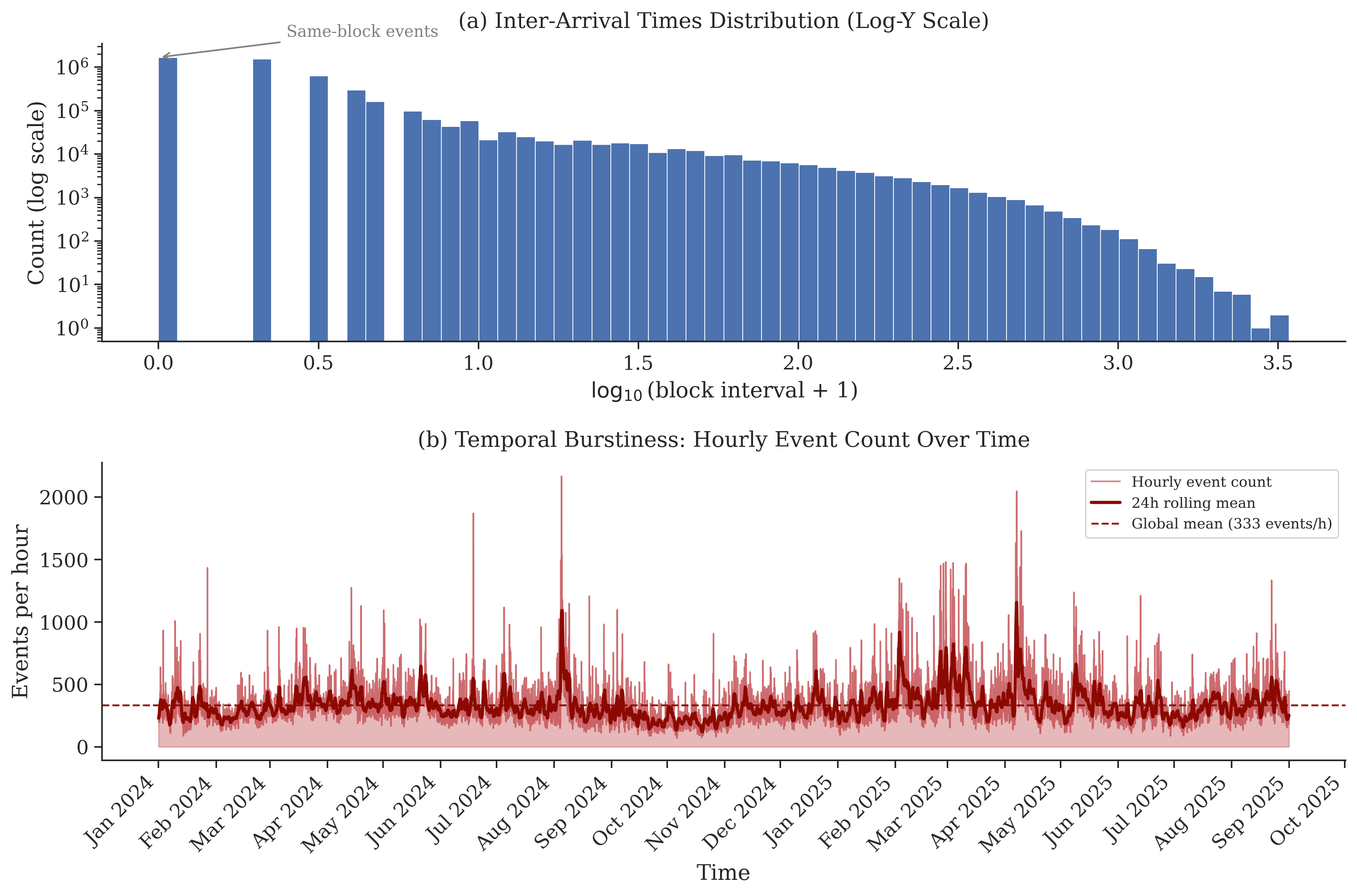}
    \caption{Uniswap Event Analysis}
    \label{fig:uniswap}
    
    \vspace{-1.em}
\end{wrapfigure}

Trading (Swap) events dominate. Across all pools, \textbf{94.2\%} of events are Swaps and \textbf{5.8\%} are Mint/Burn; the Swap-to-liquidity ratio is approximately \textbf{16.2:1} overall. The share of liquidity events varies by pool: USDC-ETH is most swap-heavy (96.0\% Swap, ratio 24.3), while WBTC-ETH and WBTC-USDC have a larger liquidity component (about 18\% Mint/Burn, ratio $\sim$4.6). This imbalance indicates that price-relevant signals (Swaps) are interspersed with liquidity updates (Mint/Burn), which can be treated as regime-changing or auxiliary events depending on the forecasting objective.

\paragraph{Temporal sparsity and irregularity.}
The data are \emph{block-indexed} and \emph{event-driven}: timestamps align with block times, and events occur only when user or protocol actions trigger them. Consecutive events are separated by a \emph{block difference} $\Delta b = \mathtt{blockNumber}_{i+1} - \mathtt{blockNumber}_i$. Our analysis yields the following:

\textbf{Block intervals:} The median inter-event block interval across all pools is \textbf{1} block; the mean is \textbf{4.5} blocks. The mean is inflated by long idle periods: per-pool averages range from 1.1 (USDC-ETH) to 31.8 (WBTC-USDT) blocks.

\textbf{Zero-event blocks:} A large fraction of blocks in the covered range contain \emph{no} Uniswap events for a given pool. For USDC-ETH, \textbf{46.4\%} of blocks are empty; for WBTC-ETH, WBTC-USDC, and WBTC-USDT, the fraction rises to \textbf{96.7--97.1\%}. Only USDC-ETH exhibits a moderate share of non-empty blocks; the others are highly sparse.

\textbf{Maximum gap:} The largest observed gap between two consecutive events within a single pool is \textbf{3{,}423} blocks (WBTC-USDT). Gaps of hundreds to thousands of blocks are common for less active pools, underscoring long periods of inactivity.

This sparsity and irregularity have direct implications for modeling. \emph{Standard continuous time-series methods} (e.g., fixed-step ARIMA, many RNN-based forecasts) assume regularly sampled observations. Here, neither clock-time nor block-time is regular at the event level: many blocks have zero events, and inter-arrival times vary from 0 (multiple events in one block) to thousands of blocks. Treating the stream as a regular series therefore requires arbitrary resampling (e.g., aggregating by block or hour), which discards event-level structure and can distort lead-lag relationships. \textbf{Event-aware} approaches that operate directly on the point process (e.g., block index and inter-arrival statistics) are better aligned with the data-generating process.

\paragraph{Burstiness.}
Hourly event counts exhibit substantial variance. The coefficient of variation (CV) of counts per hour ranges from \textbf{0.43} (USDC-ETH) to \textbf{1.18} (WBTC-USDC), indicating periods of intense activity followed by relative silence. This burstiness supports the view that AMM activity is driven by discrete, clustered events rather than a smooth, continuous flow---further motivation for event-centric forecasting.

%

\subsubsection{Lending Protocols: Aave and Morpho}

In contrast to the temporal sparsity analysis of Uniswap, we examine the \emph{market structure} and \emph{capital dynamics} of lending protocols. These protocols exhibit fundamentally different patterns: while AMMs are driven by discrete trading events, lending protocols reveal how capital flows through supply, borrowing, and repayment cycles.

\begin{wrapfigure}{r}{0.5\textwidth} 
    \centering
    \vspace{-1.em}
    
    \includegraphics[width=\linewidth]{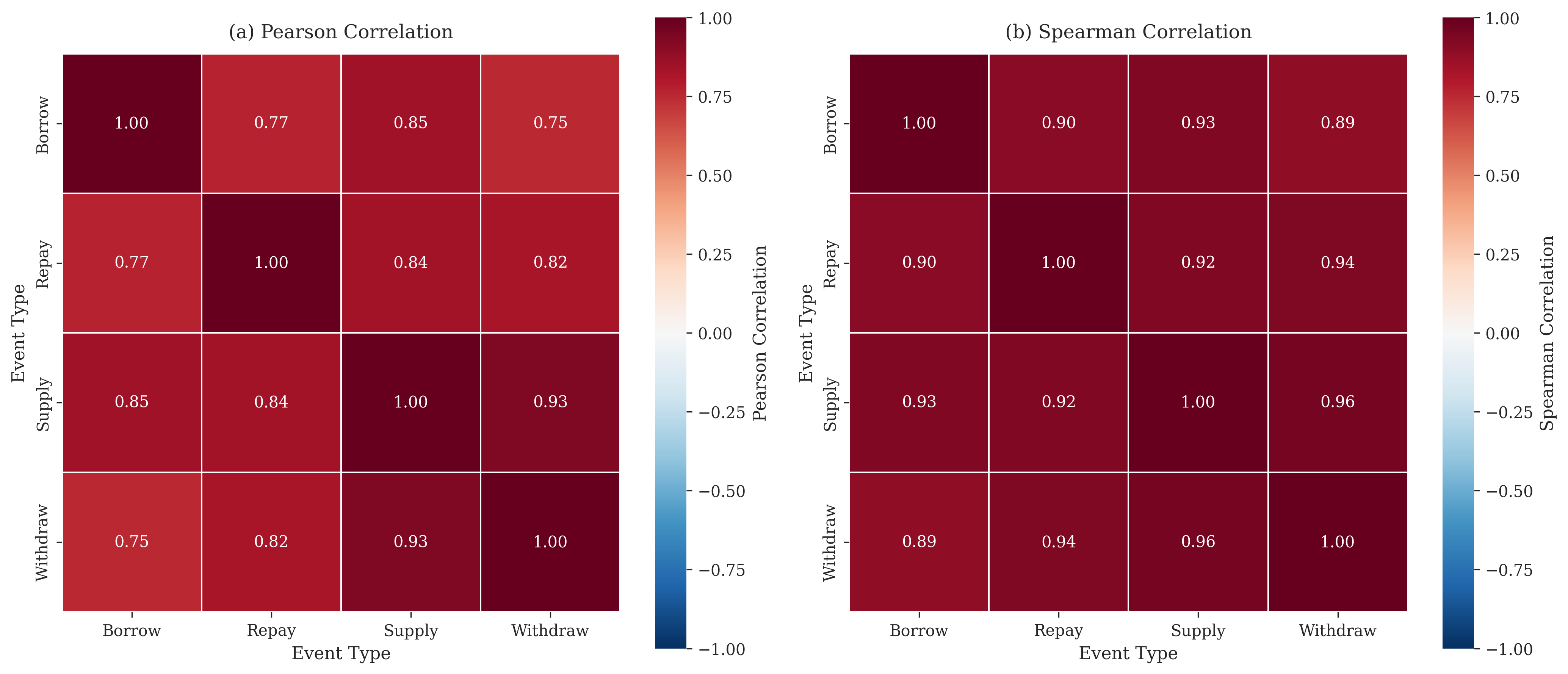}
    \caption{aave}
    \label{fig:aave}
    
    \vspace{-1.em}
\end{wrapfigure}

\begin{wrapfigure}{r}{0.5\textwidth} 
    \centering
    \vspace{-1.em}
    
    \includegraphics[width=\linewidth]{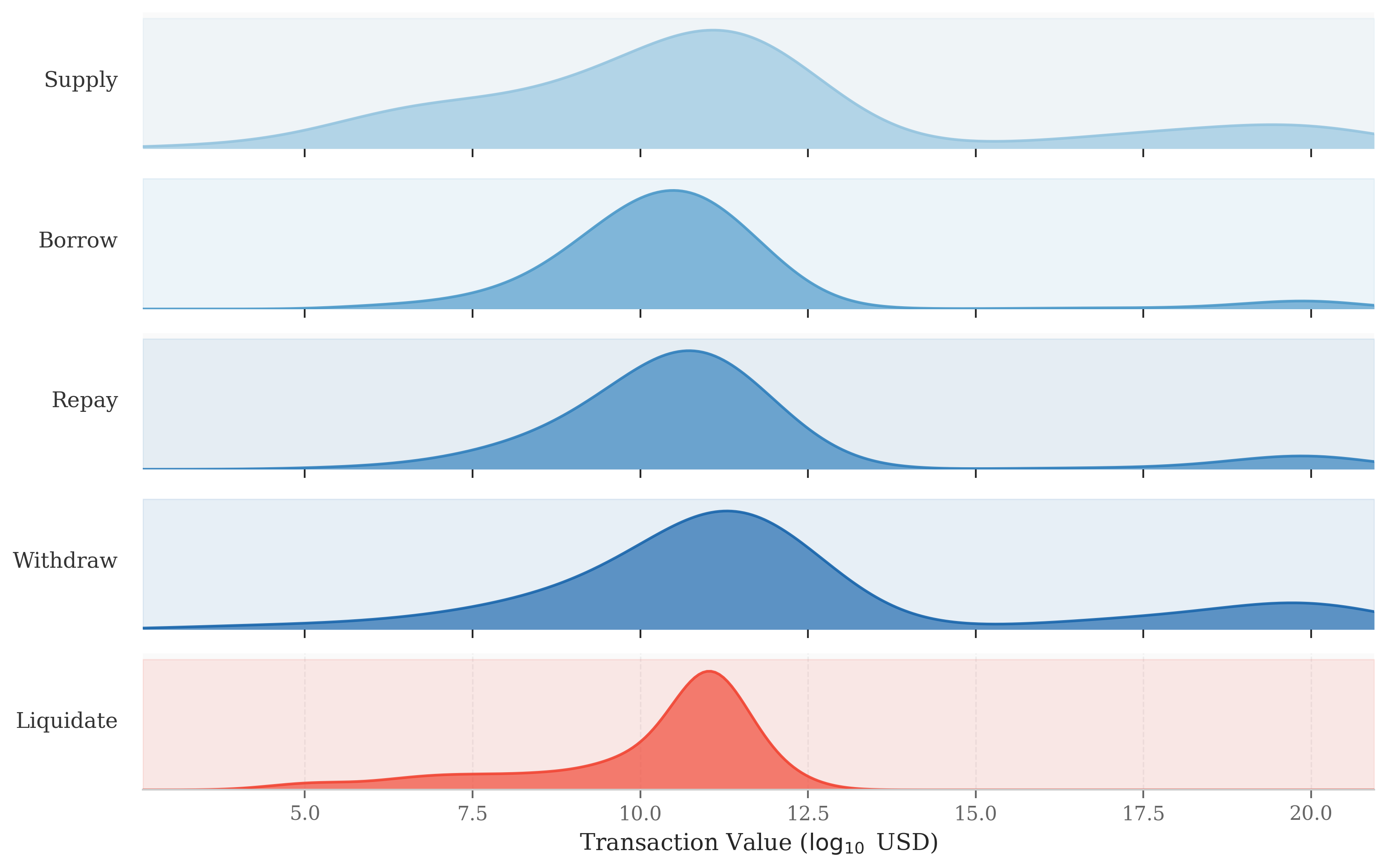}
    \caption{morpho}
    \label{fig:morpho}
    
    \vspace{-1.em}
\end{wrapfigure}

\paragraph{Value Distribution \& Whale Dominance.}
Transaction sizes in lending protocols follow heavy-tailed distributions, indicating significant concentration among large actors (``whales''). In Morpho, we analyze 94,662 events across five types: Supply (44,020), Withdraw (33,958), Borrow (10,272), Repay (6,155), and Liquidate (257). The distribution of transaction values (Figure~\ref{fig:morpho}) reveals several patterns:

\textbf{Heavy-tailed distribution:} All event types exhibit log-normal-like distributions with long tails. The median transaction sizes range from 28B (Borrow) to 253B (Withdraw) in the smallest token units, while means are inflated by extreme outliers (e.g., Supply mean: $2.82 \times 10^{19}$, median: $9.71 \times 10^{10}$).

\textbf{Event-type differences:} Withdraw events show the highest median value ($2.53 \times 10^{11}$), suggesting that users tend to withdraw larger positions than they supply. Borrow events have the lowest median ($2.80 \times 10^{10}$), indicating more frequent, smaller borrows.

\textbf{Liquidation scale:} While liquidations are rare (257 events, 0.27\% of total), they involve substantial notional value. The median liquidation size ($8.35 \times 10^{10}$) is comparable to typical borrows, but the 99th percentile reaches $2.65 \times 10^{12}$---nearly 100$\times$ the median---highlighting the risk concentration in large positions.

The heavy-tailed nature of these distributions confirms that lending protocols are dominated by a small number of large actors, with the majority of capital controlled by whales. This concentration has implications for protocol stability: large withdrawals or liquidations can cause significant market impact.

\paragraph{Event Interdependence.}
Lending protocols exhibit strong temporal coupling between different event types, reflecting the interdependent nature of supply, borrowing, and repayment cycles. We analyze Aave event counts aggregated into 4-hour windows (3,654 windows covering 2024--2025) and compute correlation matrices between Supply, Borrow, Withdraw, and Repay activities (Figure~\ref{fig:aave}). The analysis excludes \texttt{liquidationCall} events, which lack amount metadata and occur infrequently relative to core liquidity events.

The correlation analysis reveals:
\textbf{Strong supply-withdraw coupling:} Supply and Withdraw events show the highest correlation ($\rho_{\text{Pearson}} = 0.933$, $\rho_{\text{Spearman}} = 0.959$), indicating that periods of high deposit activity coincide with high withdrawal activity. This suggests active capital rotation rather than simple accumulation.

\textbf{Supply-borrow correlation:} Supply and Borrow are strongly correlated ($\rho_{\text{Pearson}} = 0.848$, $\rho_{\text{Spearman}} = 0.927$), demonstrating that available liquidity is quickly utilized. This immediate utilization pattern is consistent with efficient market behavior where capital is deployed as soon as it becomes available.

\textbf{Repayment patterns:} Repay events correlate strongly with both Supply ($\rho_{\text{Pearson}} = 0.840$) and Withdraw ($\rho_{\text{Pearson}} = 0.825$), suggesting that repayment activity often coincides with capital movements. The Spearman correlation is even higher ($\rho_{\text{Spearman}} > 0.90$ for all pairs), indicating robust rank-order relationships.

The strong temporal correlations indicate that lending protocols exhibit \emph{market coupling}: supply and borrowing activities are not independent but occur in coordinated patterns. This coupling has implications for forecasting: models that treat events as independent will miss important cross-event dependencies. Event-aware approaches that capture these interdependencies can provide more accurate predictions of protocol dynamics.

\subsubsection{Pendle}


\begin{wraptable}{l}{0.55\textwidth}
    \vspace{-15pt} 
    \centering
    \caption{Token Distribution Across Different Categories.}
    \label{Token Distribution}
    \small 
    \begin{tabular}{@{}l p{3.2cm} l@{}}
        \toprule
        \textbf{Categories} & \textbf{Token Name} & \textbf{Amt.} \\ 
        \midrule
        Stablecoin & usd, usdc, usdt, dai, frax, gho & 171 \\
        ETH & eth, weth, steth, reth, ezeth & 91 \\
        BTC & btc, wbtc, pbtc, ebtc, lbtc & 33 \\ 
        \bottomrule
    \end{tabular}
    \vspace{-10pt} 
\end{wraptable}

\textbf{Product Type and Asset Composition.} We catalogued 295 active Pendle markets as of August 2024, as described in Table \ref{Token Distribution}. The distribution reveals a significant concentration in stablecoin-based markets (58\%), followed by Ethereum liquid staking derivatives (31\%). This dominance is economically intuitive within the DeFi ecosystem: on-chain lending and restaking protocols overwhelmingly utilize interest-bearing stablecoins as collateral and reward mechanisms. Consequently, there exists a persistent, low-volatility demand for fixed-rate exposure on stable assets to hedge against yield compression. 
From a data perspective, this composition introduces a specific bias: since stablecoins are pegged to fiat, the volatility in the majority of these markets is driven almost exclusively by \textit{yield volatility} rather than \textit{underlying asset price volatility}. This separation allows our analysis to isolate pure interest-rate dynamics, distinct from the speculative noise often found in volatile crypto-asset trading.

\begin{wrapfigure}{r}{0.55\textwidth}
    \centering
    \vspace{-15pt} 
    
    \begin{subfigure}[b]{0.48\linewidth}
        \includegraphics[width=\linewidth]{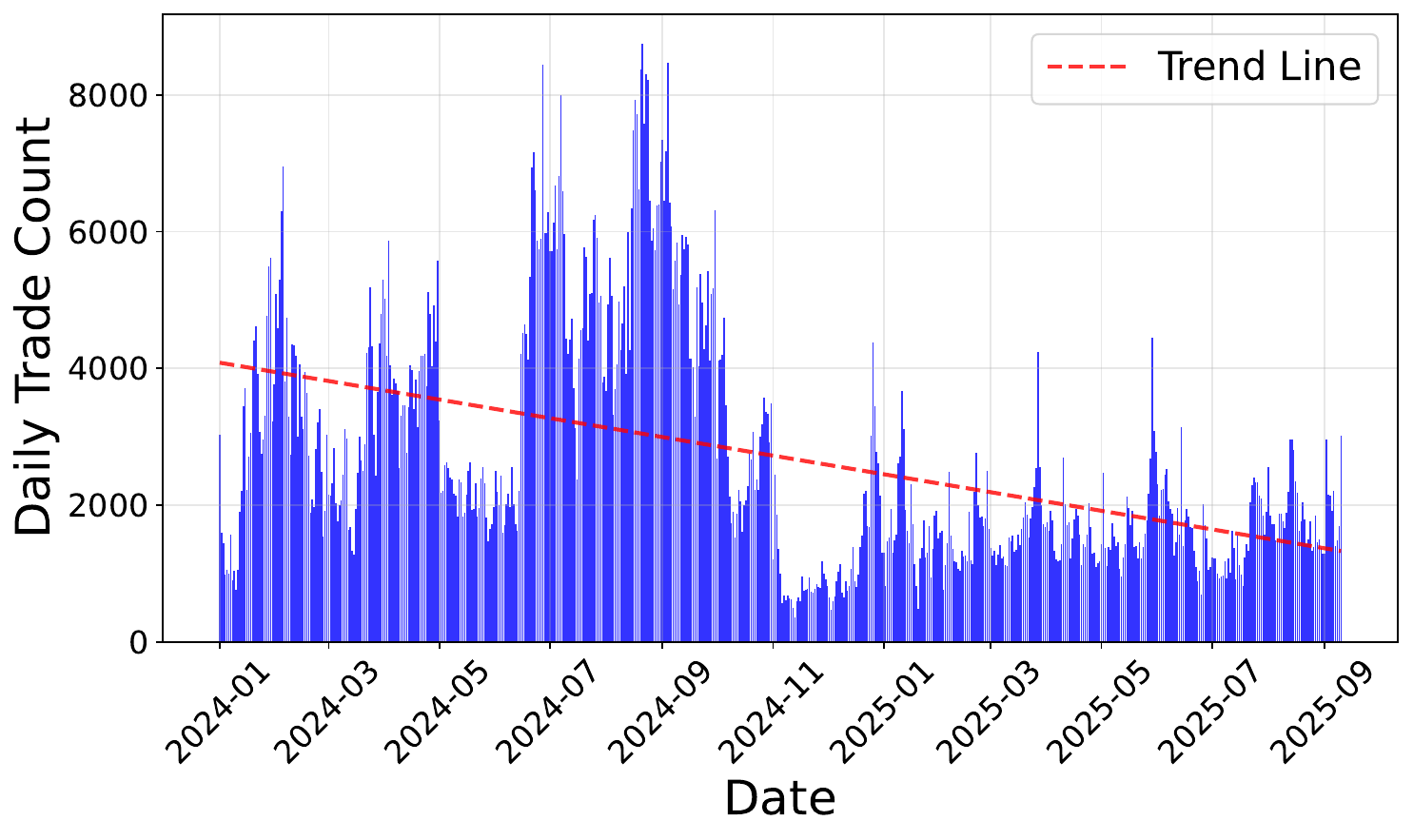}
        \caption{\small Daily Count}
        \label{fig:daily_trading_frequency}
    \end{subfigure}
    \hfill
    \begin{subfigure}[b]{0.48\linewidth}
        \includegraphics[width=\linewidth]{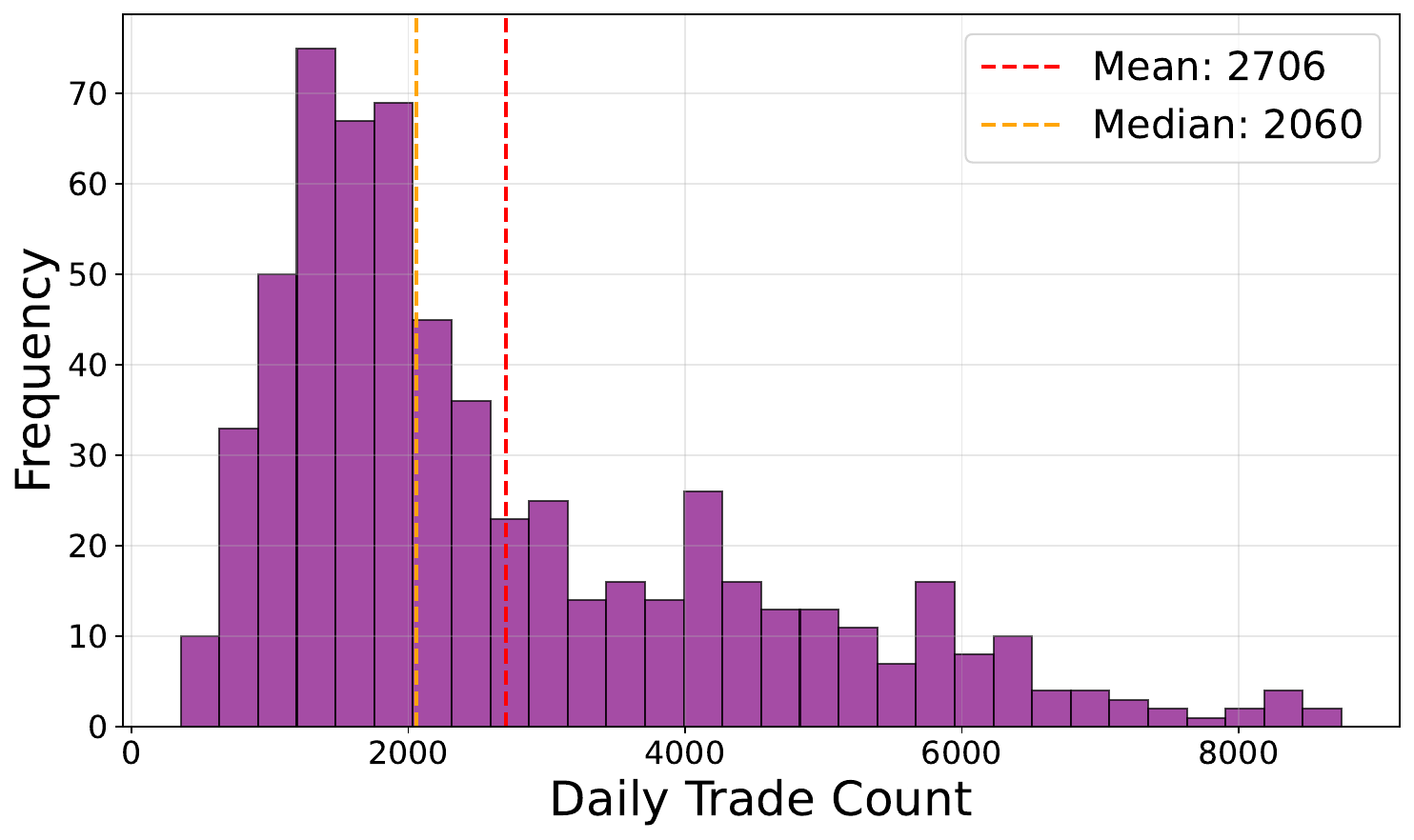}
        \caption{\small Frequency Dist.}
        \label{fig:trading_frequency_distribution}
    \end{subfigure}
    
    \par\bigskip 
    
    \begin{subfigure}[b]{0.48\linewidth}
        \includegraphics[width=\linewidth]{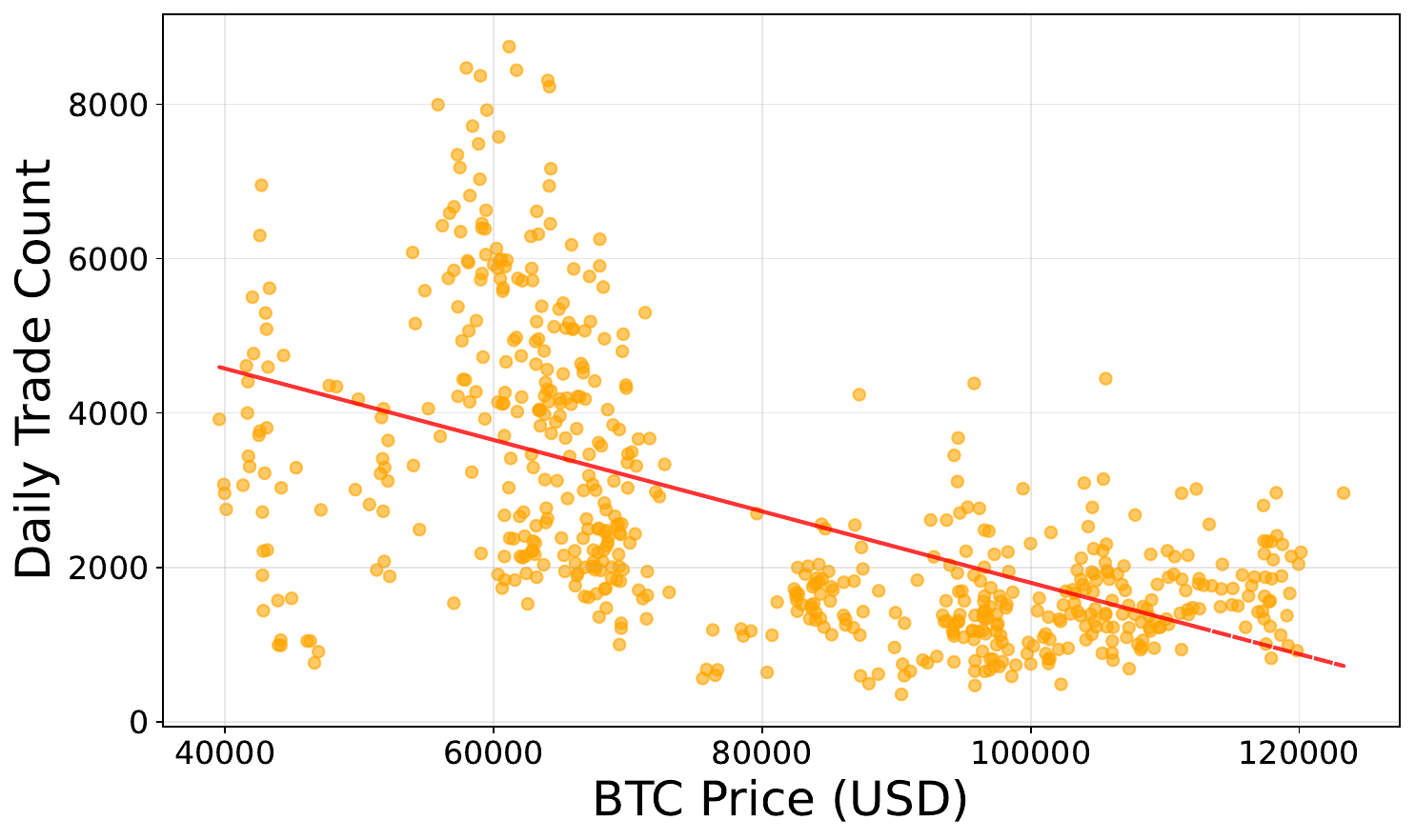}
        \caption{\small vs BTC Price}
        \label{fig:trading_frequency_vs_btc}
    \end{subfigure}
    \hfill
    \begin{subfigure}[b]{0.48\linewidth}
        \includegraphics[width=\linewidth]{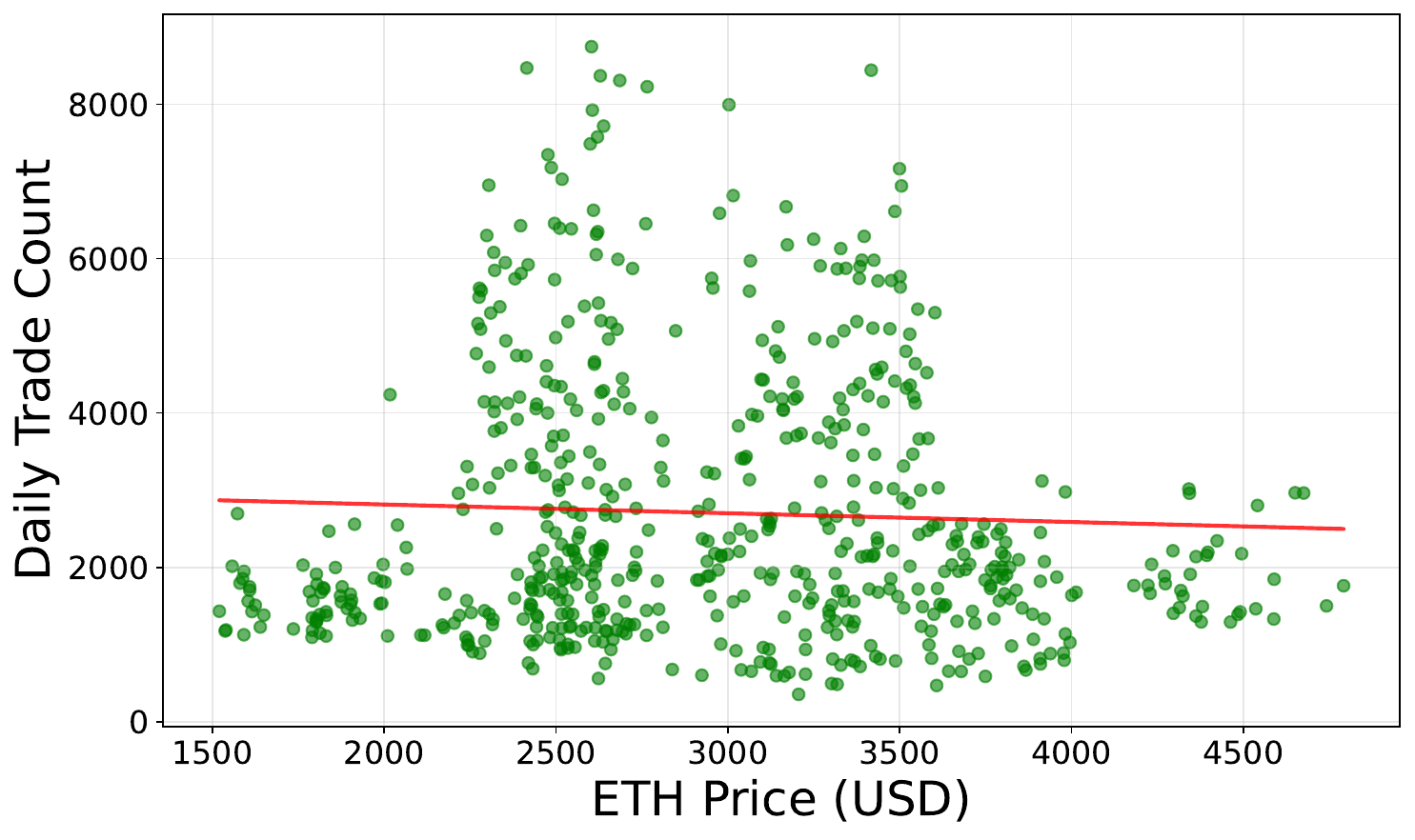}
        \caption{\small vs ETH Price}
        \label{fig:trading_frequency_vs_eth}
    \end{subfigure}
    
    \caption{Trading Frequency of Pendle Products.}
    \label{fig:trading_frequency}
    \vspace{-10pt} 
\end{wrapfigure}

\textbf{Temporal Dynamics and Market Regimes.} 
We consider each on-chain event as a discrete transaction and aggregate them daily to measure frequency. Our dataset encompasses nearly two million transactions across 295 products spanning from January 2024 to September 2025. 
The temporal analysis, presented in Figure \ref{fig:trading_frequency}, uncovers three critical behavioral patterns relevant to forecasting:

\noindent(1) \textit{Declining Trend and Market Maturation:} As shown in Figure \ref{fig:daily_trading_frequency}, the trend line exhibits a gradual downward slope. This suggests that while the protocol's Total Value Locked (TVL) may be growing, the \textit{velocity} of trading per unit of capital is decreasing, a sign of market maturation where high-frequency speculative farming gives way to longer-term holding strategies.


\noindent(2) \textit{Heavy-Tailed Activity Distribution:} Figure \ref{fig:trading_frequency_distribution} illustrates that trading frequency follows a right-skewed distribution. The mean daily trading count (2,706) significantly exceeds the median (2,060), indicating the presence of "bursty" days with exceptionally high activity. These outliers typically correspond to major DeFi events (e.g., EigenLayer airdrops or sudden gas spikes), suggesting that models must be robust to non-Gaussian noise.

\noindent(3) \textit{The Capital Rotation Effect:} Perhaps the most intriguing insight is the negative correlation between trading frequency and the spot prices of Bitcoin and Ethereum (Figures \ref{fig:trading_frequency_vs_btc} and \ref{fig:trading_frequency_vs_eth}). This counter-intuitive relationship supports a "capital rotation" hypothesis: when BTC/ETH prices surge (bull market), capital flows into spot assets for delta exposure, reducing yield-derivative trading. Conversely, in stagnant or bearish market conditions, investors flock to Pendle to lock in fixed yields or speculate on interest rates, increasing protocol activity. This implies that Pendle acts as a counter-cyclical instrument, providing valuable diversification signals for holistic market forecasting.


\textbf{Product Lifecycle and Event Stratification.} 
To fully characterize the micro-structure of these markets, we analyze four key on-chain event types: \textit{Mint}, \textit{Swap}, \textit{Burn}, and \textit{UpdateImpliedRate}. These events map directly to the creation, secondary trading, redemption, and yield revaluation phases of the Pendle lifecycle.
The data reveal a clear temporal stratification. \textit{Mint} and \textit{Swap} events occur at high frequency with smaller typical sizes, representing continuous retail participation and arbitrage bots smoothing price disparities. In stark contrast, \textit{Burn} events are sparse but involve magnitudes orders significantly larger, consistent with the expiration of large institutional positions. Finally, the extremely frequent \textit{UpdateImpliedRate} events provide a high-resolution heartbeat of the system's internal accounting.
From a modeling perspective, this heterogeneity poses a challenge for uniform time-series sampling: a standard 1-hour window might capture dozens of Swaps but miss the structural "shocks" introduced by rare Burn events. This necessitates the event-aware architecture proposed in our methodology, which can dynamically adapt to the irregular arrival times of these distinct event classes.

\newpage


\begin{table*}[t]
\centering
\small
\caption{Summary of experiment environment and hyperparameters.}
\label{tab:exp_setting}
\begin{tabularx}{\textwidth}{lYlY}
\toprule
\textbf{Category} & \textbf{Specification} & \textbf{Category} & \textbf{Specification} \\
\midrule
\textbf{Hardware} & 2 $\times$ AMD EPYC 9554 (64 cores per socket, 256 threads) & \textbf{GPU} & NVIDIA GeForce RTX 5090 (32\,GB, CUDA 13.0) \\
\textbf{Memory} & 1.1\,TiB RAM & \textbf{OS / Arch.} & Linux x86\_64, AVX512 enabled \\
\midrule
\textbf{Framework} & EasyTPP v0.1.3 (PyTorch 2.8.0+cu128) & \textbf{Precision} & FP32 (single precision) \\
\textbf{Random Seeds} & \{2019, 2020, 2021, 2022, 2023\} & \textbf{Reproducibility} & Fixed seeds, identical splits \\
\midrule
\textbf{Padding} & Right-side, pad token ID = 31 &
\textbf{Truncation} & Dynamic, max length = 300 blocks \\
\midrule
\textbf{Hidden Size} & 64 & \textbf{Dropout Rate} & 0.1 \\
\textbf{MC Samples (Integral)} & 20 per step & \textbf{Loss} & NLL or UWM \\
\midrule
\textbf{Optimizer} & Adam & \textbf{Learning Rate} & \{1e-2, 1e-3, 1e-4\} (constant) \\
\textbf{Batch Size} & 16 per GPU  & \textbf{Epochs / Stop} & 100 epochs, early stop by validation log-likelihood \\
\textbf{Gradient Accumulation} & 4 & \textbf{Scheduler} & None \\
\midrule
\textbf{Validation Metric} & Log-likelihood (for selection) & \textbf{Test Metrics} & Type-Acc, Time-RMSE, OTD (horizon = 3,5,7,9,11) \\
\textbf{Evaluation Frequency} & Every epoch & \textbf{Runs} & Mean $\pm$ std. over 5 seeds \\
\midrule
\textbf{Training Time} & $\approx$1–4 h per model on RTX 5090 & \textbf{GPU Memory} & < 20 GB usage \\
\bottomrule
\end{tabularx}
\end{table*}

\begin{table*}[htbp]
\centering
\small
\caption{Evaluation results of baseline models and their Uncertainty Weighted MSE (UWM) variants of Pendle Dataset. The first row for each model is the mean, the second row is the standard deviation. We run each experiment for 5 independent seeds and report mean $\pm$ std.}
\label{tab:abl_baselin_uwm_result}
\begin{tabularx}{\textwidth}{lYYYYYY}
\toprule
 & \multicolumn{3}{c}{Baseline (NLL)} & \multicolumn{3}{c}{With UWM Loss} \\
\cmidrule(lr){2-4}\cmidrule(lr){5-7}
Model & Type-Acc $\uparrow$ & Time RMSE $\downarrow$ & OTD $\downarrow$
      & Type-Acc $\uparrow$ & Time RMSE $\downarrow$ & OTD $\downarrow$ \\
\midrule
NHP
& 0.550 & 108.888 & 9.051 & 0.555 & \textbf{94.280} & 9.265 \\
& $\pm$0.003 & $\pm$1.84 & $\pm$0.043 & $\pm$0.003 & $\pm$0.21 & $\pm$0.048 \\
\addlinespace
RMTPP
& 0.464 & 769.139 & 9.281 & 0.480 & \textbf{95.546} & 9.248 \\
& $\pm$0.018 & $\pm$147.25 & $\pm$0.070 & $\pm$0.00 & $\pm$0.1 & $\pm$0.054 \\
\addlinespace
SAHP
& 0.547 & 93.872 & 9.155 & 0.531 & 93.983 & 9.128 \\
& $\pm$0.006 & $\pm$3.85 & $\pm$0.063 & $\pm$0.021 & $\pm$0.29 & $\pm$0.070 \\
\addlinespace
THP
& 0.219 & 958.375 & 9.733 & 0.358 & \textbf{94.498} & 9.044 \\
& $\pm$0.067 & $\pm$62.965 & $\pm$0.116 & $\pm$0.014 & $\pm$1.279 & $\pm$0.041 \\
\addlinespace
AttNHP
& 0.484 & 100.634 & 9.151 & 0.531 & \textbf{96.730} & 9.040 \\
& $\pm$0.004 & $\pm$2.222 & $\pm$0.027 & $\pm$0.003 & $\pm$3.32 & $\pm$0.032 \\
\addlinespace
IntensityFree
& 0.558 & 81.318 & 9.607 & -- & -- & -- \\
& $\pm$0.001 & $\pm$2.497 & $\pm$0.016 & -- & -- & -- \\
\addlinespace
FullyNN
& 0.471 & 665.560 & 9.762 & 0.431 & \textbf{91.084} & 9.430 \\
& $\pm$0.042 & $\pm$110.6 & $\pm$0.159 & $\pm$0.062 & $\pm$1.714 & $\pm$0.089 \\
\addlinespace
ODETPP
& 0.431 & 6815.468 & 9.820 & 0.195 & \textbf{277.069} & 9.739 \\
& $\pm$0.040 & $\pm$1266.82 & $\pm$0.056 & $\pm$0.092 & $\pm$2.24 & $\pm$0.277 \\
\bottomrule
\end{tabularx}
\end{table*}

\begin{table*}[htbp]
\centering
\small
\caption{Evaluation results of baseline models and their Uncertainty Weighted MSE (UWM) variants of Uniswap V3 Dataset. The first row for each model is the mean, the second row is the standard deviation. We run each experiment for 5 independent seeds and report mean $\pm$ std.}
\label{tab:uniswap_uwm_result}
\begin{tabularx}{\textwidth}{lYYYYYY}
\toprule
 & \multicolumn{3}{c}{Baseline (NLL)} & \multicolumn{3}{c}{With UWM Loss} \\
\cmidrule(lr){2-4}\cmidrule(lr){5-7}
Model & Type-Acc $\uparrow$ & Time RMSE $\downarrow$ & OTD $\downarrow$
      & Type-Acc $\uparrow$ & Time RMSE $\downarrow$ & OTD $\downarrow$ \\
\midrule
NHP
& 0.454 & 5.259 & 9.051 & 0.453 & \textbf{4.704} & 9.265 \\
& $\pm$0.002 & $\pm$0.124 & $\pm$0.039 & $\pm$0.002 & $\pm$0.098 & $\pm$0.042 \\
\addlinespace
RMTPP
& 0.459 & 5.283 & 9.281 & 0.428 & 6.591 & 9.248 \\
& $\pm$0.009 & $\pm$0.074 & $\pm$0.065 & $\pm$0.005 & $\pm$0.537 & $\pm$0.051 \\
\addlinespace
SAHP
& 0.405 & 4.640 & 9.155 & 0.407 & \textbf{4.630} & 9.128 \\
& $\pm$0.002 & $\pm$0.049 & $\pm$0.058 & $\pm$0.018 & $\pm$0.018 & $\pm$0.062 \\
\addlinespace
THP
& 0.462 & 5.059 & 9.733 & 0.450 & \textbf{5.056} & 9.044 \\
& $\pm$0.020 & $\pm$0.030 & $\pm$0.102 & $\pm$0.020 & $\pm$0.004 & $\pm$0.035 \\
\addlinespace
AttNHP
& 0.340 & 25.490 & 9.151 & 0.430 & \textbf{5.030} & 9.040 \\
& $\pm$0.020 & $\pm$2.310 & $\pm$0.025 & $\pm$0.010 & $\pm$0.020 & $\pm$0.029 \\
\addlinespace
IntensityFree
& 0.447 & 4.638 & 9.607 & -- & -- & -- \\
& $\pm$0.015 & $\pm$0.159 & $\pm$0.019 & -- & -- & -- \\
\addlinespace
FullyNN
& 0.074 & 108.725 & 9.762 & 0.082 & \textbf{105.43} & 9.430 \\
& $\pm$0.011 & $\pm$1.606 & $\pm$0.145 & $\pm$0.008 & $\pm$1.295 & $\pm$0.081 \\
\addlinespace
ODETPP
& 0.438 & 5.198 & 9.820 & 0.356 & 7.469 & 9.739 \\
& $\pm$0.005 & $\pm$0.285 & $\pm$0.052 & $\pm$0.093 & $\pm$1.389 & $\pm$0.215 \\
\bottomrule
\end{tabularx}
\end{table*}

\begin{table*}[t]
\centering
\caption{Event information and field descriptions in Uniswap V3.}
\label{tab:uniswap_events}
\begin{tabularx}{\textwidth}{l l X}
\toprule
\textbf{Information category} & \textbf{Name} & \textbf{Description} \\
\midrule
\multirow{4}{*}{\textbf{Basic Information}}
    & type             & The event type: one of \texttt{mint}, \texttt{burn}, or \texttt{swap} \\
\cmidrule(lr){2-3}
    & timestamp        & Unix timestamp of the block in which the transaction occurred \\
\cmidrule(lr){2-3}
    & blockNumber      & The block height at which the event occurred \\
\cmidrule(lr){2-3}
    & txHash           & The unique hash of the transaction (0x-prefixed hexadecimal string) \\
\midrule
\multirow{1}{*}{\textbf{Detailed Information By Event}}
    &                  & Parameters vary by event type; see \texttt{amount0} and \texttt{amount1} below. \\
\midrule
\multirow{2}{*}{\textbf{Mint}}
    & amount0          & Amount of token0 added to the pool; empty string if not recorded \\
\cmidrule(lr){2-3}
    & amount1          & Amount of token1 added to the pool; empty string if not recorded \\
\midrule
\multirow{2}{*}{\textbf{Burn}}
    & amount0          & Amount of token0 withdrawn from the pool; empty string if not recorded \\
\cmidrule(lr){2-3}
    & amount1          & Amount of token1 withdrawn from the pool; empty string if not recorded \\
\midrule
\multirow{2}{*}{\textbf{Swap}}
    & amount0          & Signed amount of token0: negative = outgoing from the pool, positive = incoming to the pool \\
\cmidrule(lr){2-3}
    & amount1          & Signed amount of token1: negative = outgoing from the pool, positive = incoming to the pool \\
\bottomrule
\end{tabularx}
\end{table*}


\begin{table*}[t]
\centering
\caption{Event information and field descriptions in Aave Protocol.}
\label{tab:aave_events}
\begin{tabularx}{\textwidth}{l l X}
\toprule
\textbf{Information category} & \textbf{Name} & \textbf{Description} \\
\midrule
\multirow{10}{*}{\textbf{Basic Information}}
    & event\_type       & The transaction event type: one of Supply, Borrow, Withdraw, or Repay \\
\cmidrule(lr){2-3}
    & timestamp         & The exact time when the transaction occurred (Unix timestamp of the block) \\
\cmidrule(lr){2-3}
    & block\_number     & The block height at which the transaction occurred \\
\cmidrule(lr){2-3}
    & transaction\_hash & The unique hash value of the transaction \\
\cmidrule(lr){2-3}
    & tx\_index         & The index of the transaction within the block \\
\cmidrule(lr){2-3}
    & log\_index        & The log index of this event within the transaction \\
\cmidrule(lr){2-3}
    & asset             & The address of the reserve token (0x-prefixed hexadecimal string) \\
\cmidrule(lr){2-3}
    & asset\_role       & Role of the asset, typically ``reserve'' \\
\cmidrule(lr){2-3}
    & user              & The address of the user performing the action \\
\cmidrule(lr){2-3}
    & on\_behalf\_of    & The address on whose behalf the action is performed; zero address for direct actions \\
\midrule
\multirow{1}{*}{\textbf{Detailed Information By Event}}
    &                   & Parameters vary depending on the type of event \\
\midrule
\multirow{2}{*}{\textbf{Supply}}
    & referral\_code    & Referral code used (optional; only present for some supply events) \\
\cmidrule(lr){2-3}
    & amount            & Amount of tokens supplied (optional; may be missing for some events) \\
\midrule
\multirow{1}{*}{\textbf{Borrow}}
    &                   & No additional fields beyond Basic Information \\
\midrule
\multirow{2}{*}{\textbf{Withdraw}}
    & amount            & The amount of tokens withdrawn from the reserve \\
\cmidrule(lr){2-3}
    & to                & The recipient address receiving the withdrawn tokens \\
\midrule
\multirow{3}{*}{\textbf{Repay}}
    & amount            & The amount of tokens repaid to the reserve \\
\cmidrule(lr){2-3}
    & repayer           & The address of the account repaying the debt \\
\cmidrule(lr){2-3}
    & use\_a\_tokens    & Whether repayment was made using aTokens (\texttt{true}) or the underlying asset (\texttt{false}) \\
\bottomrule
\end{tabularx}
\end{table*}


\begin{table*}[t]
\centering
\caption{Event information and field descriptions in Morpho Protocol.}
\label{tab:morpho_events}
\begin{tabularx}{\textwidth}{l l X}
\toprule
\textbf{Information category} & \textbf{Name} & \textbf{Description} \\
\midrule
\multirow{5}{*}{\textbf{Basic Information}}
    & event\_type       & The transaction event type: one of Supply, Borrow, Withdraw, Repay, or Liquidate \\
\cmidrule(lr){2-3}
    & block\_number     & The block height at which the transaction occurred \\
\cmidrule(lr){2-3}
    & transaction\_hash & The unique hash value of the transaction \\
\cmidrule(lr){2-3}
    & log\_index        & The log index of this event within the transaction \\
\cmidrule(lr){2-3}
    & market\_id        & Unique identifier for the Morpho market (0x-prefixed hexadecimal string) \\
\midrule
\multirow{1}{*}{\textbf{Detailed Information By Event}}
    &                   & All event types share the following fields; \texttt{event\_type} distinguishes the action. \\
\midrule
\multirow{4}{*}{\textbf{Supply / Borrow / Withdraw / Repay / Liquidate}}
    & user              & The address of the user performing the action (0x-prefixed hexadecimal string) \\
\cmidrule(lr){2-3}
    & receiver          & The address receiving the tokens (0x-prefixed hexadecimal string) \\
\cmidrule(lr){2-3}
    & amount            & Gross amount of tokens involved, in wei or smallest unit \\
\cmidrule(lr){2-3}
    & net\_amount       & Net amount after protocol fees and adjustments, in wei or smallest unit \\
\bottomrule
\end{tabularx}
\end{table*}

\begin{table*}[t]
\centering
\caption{Event information and field descriptions in Pendle Protocol.}
\label{tab:pendle_events}
\begin{tabularx}{\textwidth}{l l X}
\toprule
\textbf{Information category} & \textbf{Name} & \textbf{Description} \\
\midrule
\multirow{5}{*}{\textbf{Basic Information}} 
    & event\_type       & The transaction event type: one of Mint, Burn, UpdateImpliedRate, or Swap \\
\cmidrule(lr){2-3}
    & timestamp         & The exact time when the transaction occurred \\
\cmidrule(lr){2-3}
    & block\_number     & The block height at which the transaction occurred \\
\cmidrule(lr){2-3}
    & transaction\_hash & The unique hash value of the transaction \\
\cmidrule(lr){2-3}
    & log\_index        & The log index of this event within the block \\
\midrule
\multirow{1}{*}{\textbf{Detailed Information By Event}} 
    &                   & Parameters vary depending on the type of event \\
\midrule
\multirow{4}{*}{\textbf{Mint}} 
    & receiver          & The address receiving the newly minted LP tokens \\
\cmidrule(lr){2-3}
    & netLpMinted       & The actual amount of LP tokens minted and assigned to the receiver \\
\cmidrule(lr){2-3}
    & netSyUsed         & The actual amount of SY tokens used and deposited into the pool for liquidity \\
\cmidrule(lr){2-3}
    & netPtUsed         & The actual amount of PT tokens used and deposited into the pool for liquidity \\
\midrule
\multirow{4}{*}{\textbf{Burn}} 
    & receiverSy        & The address receiving the returned SY tokens \\
\cmidrule(lr){2-3}
    & receiverPt        & The address receiving the returned PT tokens \\
\cmidrule(lr){2-3}
    & netLpBurned       & The amount of LP tokens burned in this transaction \\
\cmidrule(lr){2-3}
    & netSyOut          & The actual amount of SY tokens withdrawn after burning \\
\cmidrule(lr){2-3}
    & netPtOut          & The actual amount of PT tokens withdrawn after burning \\
\midrule
\multirow{1}{*}{\textbf{UpdateImpliedRate}} 
    & lnLastImpliedRate & The implied rate at the previous block \\
\midrule
\multirow{6}{*}{\textbf{Swap}} 
    & caller            & The external contract/account initiating the call, and the actual payer of SY tokens \\
\cmidrule(lr){2-3}
    & receiver          & The target address receiving the exact amount of PT tokens \\
\cmidrule(lr){2-3}
    & netPtOut          & The exact amount of PT the user wants to receive; negative means PT flows into the pool, positive means PT flows to the receiver \\
\cmidrule(lr){2-3}
    & netSyOut          & The amount of SY the user must deposit into the pool to get PT; negative means SY flows into the pool, positive means SY flows to the receiver \\
\cmidrule(lr){2-3}
    & netSyFee          & The SY fee paid to liquidity providers, retained in the pool to increase reserves \\
\cmidrule(lr){2-3}
    & netSyToReserve    & Additional SY transferred to the protocol reserve for long-term incentives and insurance \\
\bottomrule
\end{tabularx}
\end{table*}

\newpage
\section{Complete Evaluation Results of Baseline Models and UWM Variants}
\label{results}

\textcolor{red}{
Table~\ref{tab:abl_baselin_uwm_result}  and Table ~\ref{tab:uniswap_uwm_result} present the full evaluation results of the baseline models and their Uncertainty Weighted MSE (UWM) variants in Pendle dataset and Uniswap V3 dataset.. 
A partial version of this table was included in the table \ref{tab:combined_uwm_results}. Here, we provide the complete set of metrics and statistics for all models. 
Higher $\uparrow$ values indicate better Type-Acc performance, while lower $\downarrow$ values indicate better RMSE and OTD. 
}

\newpage
\section{Experiment Settings}
\label{sec:appendixC}

\noindent
To ensure reproducibility, all experiments were performed under a consistent hardware-software environment. 
Table~\ref{tab:exp_setting} summarizes the complete setup, covering hardware, framework, model configuration, and key hyperparameters.

\end{document}